\documentclass{article}

\usepackage[main, preprint]{neurips_2026}

\usepackage[utf8]{inputenc}
\usepackage[T1]{fontenc}
\usepackage{url}
\usepackage{booktabs}
\usepackage{graphicx}
\usepackage{amsmath}
\usepackage{amssymb}
\usepackage{nicefrac}
\usepackage{microtype}
\usepackage{xcolor}

\usepackage{wrapfig}
\usepackage{multirow}
\usepackage[table]{xcolor}

\usepackage{pgfplots}
\usepackage{subcaption}
\usepackage{booktabs}
\pgfplotsset{compat=1.18}
\usepackage{rotating}
\usepackage{multirow}
\usepackage{algorithm}
\usepackage{algorithmic}
\usepackage{pifont}

\definecolor{FeatCalRow}{RGB}{235,245,255}
\newcommand{\method}{E-PMQ}
\newcommand{\zerogain}{\textcolor{gray}{\tiny$\uparrow$0.0}}
\newcommand{\upgain}[1]{\textcolor{gray}{\tiny$\uparrow$#1}}
\newcommand{\downgain}[1]{\textcolor{gray}{\tiny$\downarrow$#1}}

\definecolor{CodeGreen}{HTML}{8EE8AA}
\usepackage{hyperref}
\author{%
Wenjun Wang$^{1}${\thanks{Equal contribution.}} \quad
Yanggan Gu$^{1*}$ \quad
Shuo Cai$^{1*}$ \quad
Yuanyi Wang$^{1}$  \\ 
\textbf{Pengkai Wang$^{1}$}  \quad
\textbf{Jianmin Wu$^{1,2}$} \quad
\textbf{Hongxia Yang$^{1,2,3}$}\thanks{Corresponding author: \texttt{hongxia.yang@polyu.edu.hk}.} \\
$^{1}$The Hong Kong Polytechnic University \\
$^{2}$ PolyU-Daya Bay Technology and Innovation Research Institute \\
$^{3}$InfiX.ai \\
wenjun369.wang@connect.polyu.hk \\
\textbf{Code:}
\href{https://github.com/wwjzhy/E-PMQ}{\underline{\texttt{github.com/wwjzhy/E-PMQ}}}
}

\begin{document}

\title{E-PMQ: Expert-Guided Post-Merge Quantization with Merged-Weight Anchoring}
\maketitle

% Auto-generated from 粘贴的文本 (1)(7).txt
% Load this file in the preamble with: \input{tables_and_figures_commands}
% Then call commands such as \figMain or \tabClipVitBThirtyTwoPmqResults at the desired locations.

\newcommand{\figMain}{%
\begin{figure}[t]
    \centering
    \includegraphics[width=\linewidth]{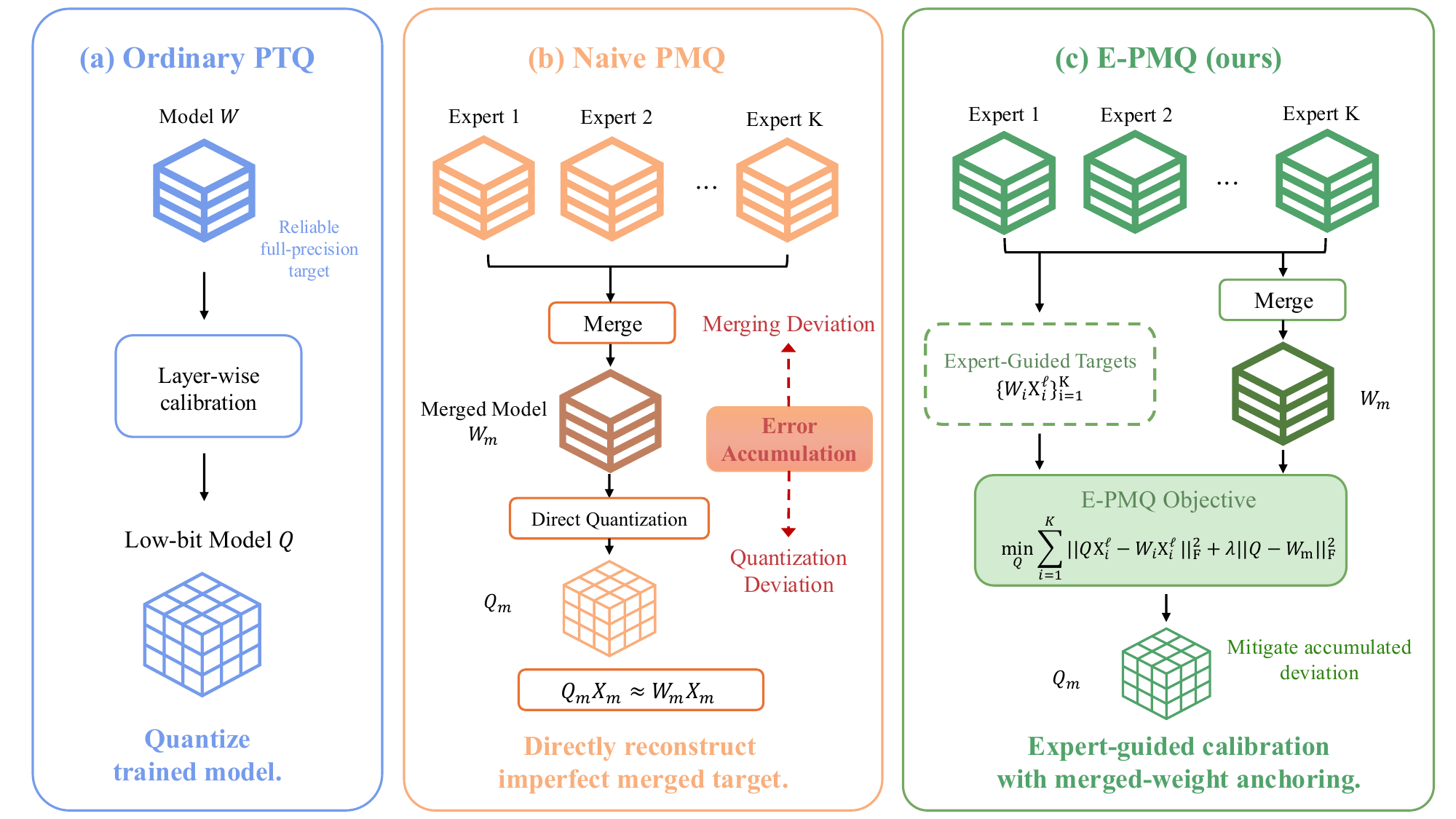}
    \caption{
    Overview of ordinary PTQ, naive PMQ, and E-PMQ.
    Ordinary PTQ quantizes a trained model by reconstructing a reliable full-precision target.
    Naive PMQ first merges multiple expert checkpoints and then directly quantizes the merged model, thereby reconstructing an imperfect merged target and suffering from accumulated merging and quantization deviation.
    E-PMQ instead uses expert-guided output targets during layer-wise calibration and anchors the quantized weight to the merged checkpoint $W_m$, turning post-merge quantization into expert-guided calibration with merged-weight anchoring.
    }
    \label{fig:main}
    \vspace{-2em}
\end{figure}
}

\newcommand{\tabClipVitBThirtyTwoPmqResults}{%
\begin{table}[t]
\centering
\caption{
Performance of 4-bit post-merge quantized CLIP-ViT-B/32 models on eight image-classification tasks.
All numbers are top-1 accuracy (\%). Gray arrows show changes over the corresponding full-precision merged checkpoints.
}
\label{tab:clip-vit-b32-pmq-results}
\setlength{\tabcolsep}{3.2pt}
\resizebox{\textwidth}{!}{
\begin{tabular}{l|cccccccc|c}
\toprule
Method & SUN397 & Cars & RESISC45 & EuroSAT & SVHN & GTSRB & MNIST & DTD & Avg. \\
\midrule
Pre-trained 
& 63.2 & 59.8 & 60.7 & 46.0 & 31.6 & 32.5 & 48.2 & 43.9 & 48.2 \\
Fine-tuned (STL) 
& 75.0 & 78.3 & 95.2 & 99.0 & 97.3 & 98.9 & 99.6 & 79.7 & 90.3 \\
\midrule
Simple Averaging 
& 65.4\zerogain & 62.6\zerogain & 70.8\zerogain & 76.9\zerogain & 64.5\zerogain & 54.9\zerogain & 86.3\zerogain & 50.9\zerogain & 66.5\zerogain \\

\quad w/ RTN 
& 62.0\downgain{3.4} & 54.7\downgain{7.9} & 65.6\downgain{5.2} & 60.2\downgain{16.7} & 61.6\downgain{2.9} & 47.7\downgain{7.2} & 84.4\downgain{1.9} & 46.3\downgain{4.6} & 60.3\downgain{6.2} \\

\quad w/ GPTQ 
& 64.3\downgain{1.2} & 59.5\downgain{3.1} & 69.3\downgain{1.5} & 68.0\downgain{8.9} & 64.0\downgain{0.5} & 49.1\downgain{5.8} & 86.4\upgain{0.1} & 49.4\downgain{1.5} & 63.7\downgain{2.8} \\

\quad w/ AWQ 
& 61.7\downgain{3.7} & 51.7\downgain{10.9} & 64.5\downgain{6.3} & 57.8\downgain{19.1} & 63.8\downgain{0.7} & 49.6\downgain{5.3} & 84.3\downgain{2.0} & 46.7\downgain{4.2} & 60.0\downgain{6.5} \\

\rowcolor{FeatCalRow}\quad w/ \method 
& 67.7\upgain{2.3} & 67.8\upgain{5.2} & 80.5\upgain{9.7} & 68.6\downgain{8.3} & 94.9\upgain{30.4} & 59.5\upgain{4.6} & 99.0\upgain{12.7} & 63.1\upgain{12.2} & \textbf{75.1}\upgain{8.6} \\
\midrule
Task Arithmetic ($\lambda=0.3$) 
& 57.1\zerogain & 55.7\zerogain & 64.9\zerogain & 76.7\zerogain & 77.9\zerogain & 68.5\zerogain & 96.1\zerogain & 47.2\zerogain & 68.0\zerogain \\

\quad w/ RTN 
& 52.0\downgain{5.1} & 45.7\downgain{10.0} & 59.9\downgain{5.0} & 62.8\downgain{13.9} & 75.4\downgain{2.5} & 54.6\downgain{13.9} & 95.1\downgain{1.0} & 44.3\downgain{2.9} & 61.2\downgain{6.8} \\

\quad w/ GPTQ 
& 55.6\downgain{1.5} & 53.5\downgain{2.2} & 63.8\downgain{1.1} & 69.1\downgain{7.6} & 77.9\zerogain & 57.0\downgain{11.5} & 95.9\downgain{0.2} & 47.5\upgain{0.3} & 65.0\downgain{3.0} \\

\quad w/ AWQ 
& 52.2\downgain{4.9} & 46.3\downgain{9.4} & 60.3\downgain{4.6} & 58.6\downgain{18.1} & 74.9\downgain{3.1} & 55.6\downgain{12.9} & 95.3\downgain{0.8} & 43.7\downgain{3.5} & 60.9\downgain{7.1} \\

\rowcolor{FeatCalRow}\quad w/ \method 
& 67.0\upgain{9.9} & 64.4\upgain{8.7} & 78.5\upgain{13.6} & 66.3\downgain{10.4} & 94.8\upgain{16.9} & 57.0\downgain{11.5} & 98.9\upgain{2.8} & 62.0\upgain{14.8} & \textbf{73.6}\upgain{5.6} \\
\midrule
TIES-Merging ($\lambda=0.3$) 
& 67.1\zerogain & 64.2\zerogain & 74.1\zerogain & 76.8\zerogain & 77.7\zerogain & 69.4\zerogain & 94.1\zerogain & 54.0\zerogain & 72.2\zerogain \\

\quad w/ RTN
& 63.1\downgain{4.0} & 55.0\downgain{9.2} & 69.3\downgain{4.8} & 58.9\downgain{17.9} & 75.2\downgain{2.5} & 59.8\downgain{9.6} & 93.1\downgain{1.0} & 48.7\downgain{5.3} & 65.4\downgain{6.8} \\

\quad w/ GPTQ 
& 65.6\downgain{1.5} & 61.1\downgain{3.1} & 72.3\downgain{1.8} & 67.7\downgain{9.1} & 76.7\downgain{1.0} & 62.9\downgain{6.5} & 93.9\downgain{0.2} & 53.0\downgain{1.0} & 69.1\downgain{3.1} \\

\quad w/ AWQ 
& 62.9\downgain{4.2} & 55.4\downgain{8.8} & 68.8\downgain{5.3} & 56.6\downgain{20.2} & 76.2\downgain{1.5} & 59.3\downgain{10.1} & 93.3\downgain{0.8} & 49.3\downgain{4.7} & 65.2\downgain{7.0} \\

\rowcolor{FeatCalRow}\quad w/ \method 
& 67.6\upgain{0.5} & 66.7\upgain{2.5} & 80.5\upgain{6.4} & 67.2\downgain{9.6} & 94.7\upgain{17.0} & 59.2\downgain{10.2} & 99.0\upgain{4.9} & 63.2\upgain{9.2} & \textbf{74.8}\upgain{2.6} \\
\midrule
WUDI-Merging 
& 68.0\zerogain & 72.5\zerogain & 85.0\zerogain & 94.6\zerogain & 94.8\zerogain & 94.9\zerogain & 99.3\zerogain & 66.6\zerogain & \textbf{84.5}\zerogain \\

\quad w/ RTN
& 62.9\downgain{5.1} & 63.8\downgain{8.7} & 79.4\downgain{5.6} & 85.5\downgain{9.1} & 94.2\downgain{0.6} & 80.0\downgain{14.9} & 99.1\downgain{0.2} & 59.6\downgain{7.0} & 78.1\downgain{6.4} \\

\quad w/ GPTQ 
& 66.6\downgain{1.4} & 68.9\downgain{3.6} & 83.8\downgain{1.2} & 90.0\downgain{4.6} & 94.8\zerogain & 81.3\downgain{13.6} & 99.3\zerogain & 64.2\downgain{2.4} & 81.1\downgain{3.4} \\

\quad w/ AWQ 
& 62.5\downgain{5.5} & 63.8\downgain{8.8} & 79.7\downgain{5.3} & 83.9\downgain{10.7} & 93.9\downgain{0.9} & 80.2\downgain{14.7} & 99.1\downgain{0.2} & 59.7\downgain{6.9} & 77.8\downgain{6.7} \\

\rowcolor{FeatCalRow}\quad w/ \method 
& 67.9\downgain{0.1} & 68.9\downgain{3.6} & 86.8\upgain{1.8} & 92.9\downgain{1.7} & 95.3\upgain{0.5} & 80.6\downgain{14.4} & 99.3\zerogain & 68.0\upgain{1.4} & 82.4\downgain{2.1} \\
\bottomrule
\end{tabular}
}
\end{table}
}

\newcommand{\tabClipExtendedAvg}{%
\begin{wraptable}[14]{r}{0.60\columnwidth}
\centering
\vspace{-2em}
\caption{\small Extended CLIP average accuracy under 4-bit PMQ.}
\label{tab:clip-extended-avg}
\vspace{0.2em}
\setlength{\tabcolsep}{2.0pt}
\renewcommand{\arraystretch}{0.90}
\scriptsize
\resizebox{\linewidth}{!}{
\begin{tabular}{l|ccccc}
\toprule
\multirow{2}{*}{Method} 
& 8 Task & 14 Task & 14 Task & 20 Task & 20 Task \\
& L/14 & B/32 & L/14 & B/32 & L/14 \\
\midrule
Pre-trained 
& 64.6 & 58.8 & 69.1 & 55.6 & 65.6 \\
Fine-tuned 
& 94.3 & 90.0 & 92.8 & 90.3 & 93.1 \\
\midrule
Task Arithmetic 
& 80.7\zerogain & 52.8\zerogain & 63.1\zerogain & 36.3\zerogain & 57.2\zerogain \\
\quad w/ RTN 
& 77.3\downgain{3.4} & 47.6\downgain{5.2} & 57.8\downgain{5.3} & 33.1\downgain{3.2} & 33.7\downgain{23.5} \\
\quad w/ GPTQ 
& 78.3\downgain{2.4} & 49.9\downgain{2.9} & 59.2\downgain{3.9} & 35.0\downgain{1.3} & 34.8\downgain{22.4} \\
\quad w/ AWQ 
& 66.9\downgain{13.8} & 48.3\downgain{4.5} & 49.1\downgain{14.1} & 34.0\downgain{2.3} & 27.7\downgain{29.5} \\
\rowcolor{FeatCalRow}\quad w/ \method 
& \textbf{85.9}\upgain{5.2} & \textbf{70.1}\upgain{17.3} & \textbf{82.0}\upgain{18.9} & \textbf{64.2}\upgain{27.9} & \textbf{76.7}\upgain{19.5} \\
\midrule
TIES-Merging 
& 84.0\zerogain & 67.6\zerogain & 77.8\zerogain & 55.6\zerogain & 63.0\zerogain \\
\quad w/ RTN 
& 81.2\downgain{2.8} & 60.8\downgain{6.8} & 72.7\downgain{5.1} & 51.0\downgain{4.6} & 60.3\downgain{2.7} \\
\quad w/ GPTQ 
& 81.8\downgain{2.2} & 63.5\downgain{4.1} & 74.0\downgain{3.8} & 53.1\downgain{2.5} & 61.1\downgain{1.9} \\
\quad w/ AWQ 
& 73.4\downgain{10.6} & 60.8\downgain{6.8} & 67.9\downgain{9.9} & 51.4\downgain{4.2} & 58.0\downgain{5.0} \\
\rowcolor{FeatCalRow}\quad w/ \method 
& \textbf{86.0}\upgain{2.0} & \textbf{72.1}\upgain{4.5} & \textbf{82.5}\upgain{4.7} & \textbf{67.8}\upgain{12.2} & \textbf{77.5}\upgain{14.5} \\
\bottomrule
\end{tabular}
}
\vspace{-0.8em}
\end{wraptable}
}

\newcommand{\tabFlanTFiveGlueResults}{%
\begin{table*}[t]
\centering
\caption{
Main results on FLAN-T5 under 4-bit post-merge quantization on GLUE tasks.
All numbers are task scores (\%), and STS-B reports Spearman's correlation.
Gray arrows show changes over the corresponding full-precision merged checkpoints.
}
\label{tab:flan-t5-glue-results}
\setlength{\tabcolsep}{3.2pt}
\renewcommand{\arraystretch}{0.96}
\resizebox{\textwidth}{!}{
\begin{tabular}{l|cccccccc|c}
\toprule
Method & CoLA & MNLI & MRPC & QNLI & QQP & RTE & SST-2 & STS-B & Avg. \\
\midrule
Task Arithmetic 
& 69.80\zerogain & 57.66\zerogain & 78.43\zerogain & 90.26\zerogain & 83.61\zerogain & 80.51\zerogain & 92.20\zerogain & 77.82\zerogain & 78.79\zerogain \\

\quad w/ RTN 
& 70.18\upgain{0.4} & 56.29\downgain{1.4} & 78.92\upgain{0.5} & 90.04\downgain{0.2} & 83.78\upgain{0.2} & 81.59\upgain{1.1} & 91.17\downgain{1.0} & 76.10\downgain{1.7} & 78.51\downgain{0.3} \\

\quad w/ GPTQ 
& 69.42\downgain{0.4} & 55.99\downgain{1.7} & 77.70\downgain{0.7} & 89.90\downgain{0.4} & 83.73\upgain{0.1} & 79.42\downgain{1.1} & 92.20\zerogain & 77.74\downgain{0.1} & 78.26\downgain{0.5} \\

\quad w/ AWQ 
& 69.51\downgain{0.3} & 61.30\upgain{3.6} & 78.43\zerogain & 89.84\downgain{0.4} & 82.91\downgain{0.7} & 80.51\zerogain & 91.06\downgain{1.1} & 76.85\downgain{1.0} & 78.80\upgain{0.0} \\

\rowcolor{FeatCalRow}\quad w/ \method 
& 69.80\zerogain & 82.50\upgain{24.8} & 79.41\upgain{1.0} & 90.43\upgain{0.2} & 84.34\upgain{0.7} & 82.67\upgain{2.2} & 92.78\upgain{0.6} & 84.80\upgain{7.0} & \textbf{83.34}\upgain{4.6} \\
\midrule

TIES-Merging 
& 70.37\zerogain & 65.02\zerogain & 78.68\zerogain & 90.24\zerogain & 83.53\zerogain & 81.59\zerogain & 91.86\zerogain & 78.58\zerogain & 79.98\zerogain \\

\quad w/ RTN 
& 70.37\zerogain & 63.57\downgain{1.5} & 79.66\upgain{1.0} & 89.91\downgain{0.3} & 83.56\upgain{0.0} & 82.67\upgain{1.1} & 91.06\downgain{0.8} & 75.42\downgain{3.2} & 79.53\downgain{0.5} \\

\quad w/ GPTQ 
& 69.80\downgain{0.6} & 65.87\upgain{0.9} & 78.19\downgain{0.5} & 89.95\downgain{0.3} & 83.20\downgain{0.3} & 80.51\downgain{1.1} & 91.28\downgain{0.6} & 78.45\downgain{0.1} & 79.66\downgain{0.3} \\

\quad w/ AWQ 
& 68.55\downgain{1.8} & 68.16\upgain{3.1} & 78.43\downgain{0.3} & 89.91\downgain{0.3} & 82.81\downgain{0.7} & 79.78\downgain{1.8} & 90.83\downgain{1.0} & 76.51\downgain{2.1} & 79.37\downgain{0.6} \\

\rowcolor{FeatCalRow}\quad w/ \method 
& 69.51\downgain{0.9} & 82.48\upgain{17.5} & 81.37\upgain{2.7} & 90.81\upgain{0.6} & 84.30\upgain{0.8} & 80.51\downgain{1.1} & 93.00\upgain{1.1} & 85.84\upgain{7.3} & \textbf{83.48}\upgain{3.5} \\
\bottomrule
\end{tabular}
}
\vspace{-1.5em}
\end{table*}
}

\newcommand{\tabLlamaPmqResults}{%
\begin{table}[b]
\centering
\vspace{-2em}
\caption{
Performance of 4-bit post-merge quantized Llama-3.1 models merged by Task Arithmetic.
All numbers are scores (\%). Gray arrows show changes over the corresponding
full-precision merged models.
}
\label{tab:llama-pmq-results}
\setlength{\tabcolsep}{3.2pt}
\resizebox{\textwidth}{!}{
\begin{tabular}{l|cccccc|c}
\toprule
\multicolumn{8}{l}{\textit{Llama-3.1-3B}} \\
\midrule
Method & GSM8K & MATH500 & ARC-C & IFEval & HumanEval & MBPP+ & Avg. \\
\midrule
Task Arithmetic
& 74.91\zerogain & 43.20\zerogain & 72.70\zerogain & 62.48\zerogain & 53.66\zerogain & 57.94\zerogain & 60.81\zerogain \\

\quad w/ AWQ
& 74.45\downgain{0.46} & 39.60\downgain{3.60} & 72.44\downgain{0.26} & 60.63\downgain{1.85} & 51.22\downgain{2.44} & 56.08\downgain{1.86} & 59.07\downgain{1.74} \\

\quad w/ GPTQ
& 73.77\downgain{1.14} & 41.40\downgain{1.80} & 70.90\downgain{1.80} & 58.78\downgain{3.70} & 51.83\downgain{1.83} & 55.56\downgain{2.38} & 58.71\downgain{2.10} \\

\rowcolor{FeatCalRow}
\quad w/ \method
& 74.60\downgain{0.31} & 44.60\upgain{1.40} & 70.99\downgain{1.71} & 62.48\zerogain & 51.83\downgain{1.83} & 57.14\downgain{0.80} & \textbf{60.27}\downgain{0.54} \\

\midrule
\multicolumn{8}{l}{\textit{Llama-3.1-8B}} \\
\midrule
Method & GSM8K & MATH500 & ARC-C & IFEval & HumanEval & MBPP+ & Avg. \\
\midrule
Task Arithmetic
& 85.67\zerogain & 48.20\zerogain & 77.99\zerogain & 50.09\zerogain & 65.24\zerogain & 61.38\zerogain & 64.76\zerogain \\

\quad w/ AWQ
& 84.31\downgain{1.36} & 48.60\upgain{0.40} & 76.62\downgain{1.37} & 44.73\downgain{5.36} & 58.54\downgain{6.70} & 60.85\downgain{0.53} & 62.27\downgain{2.49} \\

\quad w/ GPTQ
& 85.14\downgain{0.53} & 47.00\downgain{1.20} & 76.45\downgain{1.54} & 44.92\downgain{5.17} & 58.54\downgain{6.70} & 57.94\downgain{3.44} & 61.66\downgain{3.10} \\

\rowcolor{FeatCalRow}
\quad w/ \method
& 84.23\downgain{1.44} & 45.80\downgain{2.40} & 76.28\downgain{1.71} & 48.80\downgain{1.29} & 60.98\downgain{4.26} & 61.38\zerogain & \textbf{62.91}\downgain{1.85} \\
\bottomrule
\end{tabular}
}
\end{table}
}

\newcommand{\figAnchorAblation}{
\begin{figure}[b]
    \centering
    \vspace{-1em}
    \captionsetup[subfigure]{justification=centering, skip=0pt}

    \begin{subfigure}[c]{0.32\textwidth}
        \centering
        \includegraphics[
            width=\linewidth,
            trim=0 5pt 0 10pt,
            clip
        ]{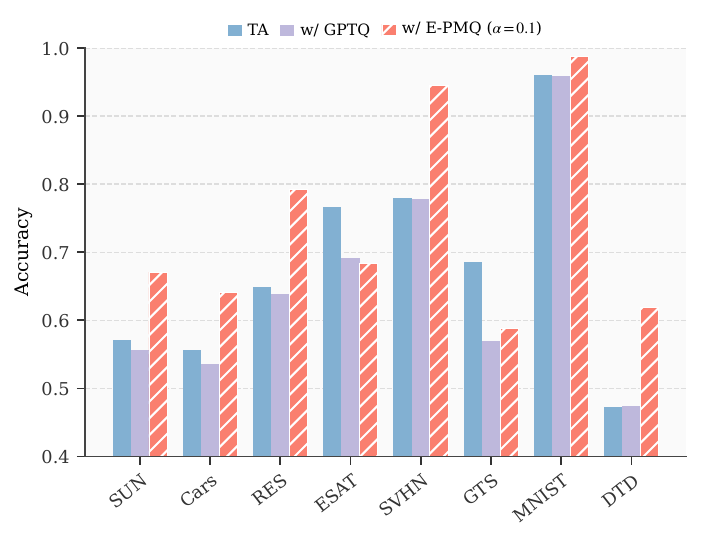}
        \caption{}
        \label{fig:ta_task_ablation}
        \vspace{-0.7em}
    \end{subfigure}
    \hfill
    \begin{subfigure}[c]{0.32\textwidth}
        \centering
        \includegraphics[
            width=\linewidth,
            trim=0 5pt 0 10pt,
            clip
        ]{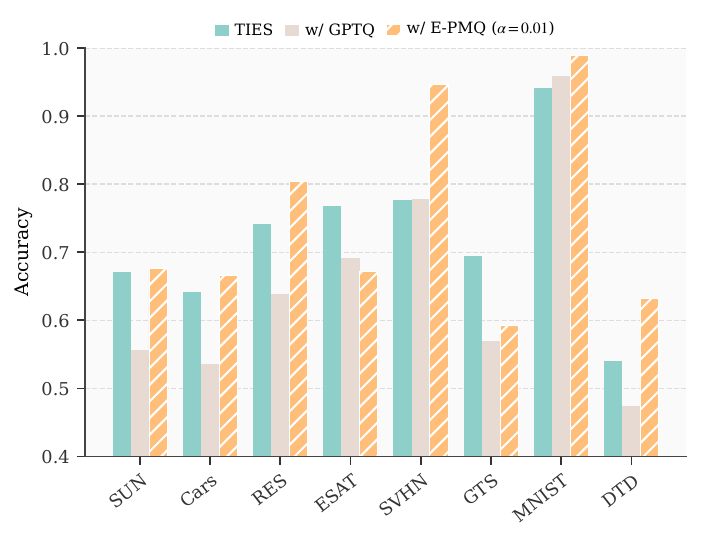}
        \caption{}
        \label{fig:ties_task_ablation}
        \vspace{-0.7em}
    \end{subfigure}
    \hfill
    \begin{subfigure}[c]{0.32\textwidth}
        \centering
        \includegraphics[
            width=\linewidth,
            trim=0 8pt 0 10pt,
            clip
        ]{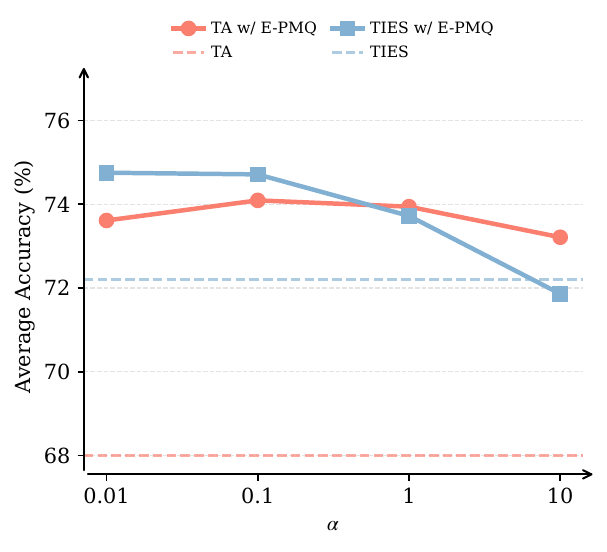}
        \caption{}
        \label{fig:alpha_sensitivity}
        \vspace{1em}
    \end{subfigure}

    \vspace{-0.8em}
    \caption{
    \textbf{Ablation analysis of \method.}
    (a) Task-level performance under Task Arithmetic (TA).
    (b) Task-level performance under TIES-Merging.
    (c) Sensitivity to the positive anchor strength $\alpha$.
    Across merging strategies, \method{} improves post-merge quantization over direct GPTQ, and its performance remains stable across a reasonable range of anchor strengths.
    }
    \label{fig:ablation_analysis}
\end{figure}
}

\newcommand{\tabAnchorAblation}{
\begin{wraptable}[10]{r}{0.3\textwidth}
    \centering
    \setlength{\tabcolsep}{5pt}
    \renewcommand{\arraystretch}{1.05}
    \resizebox{\linewidth}{!}{
    \begin{tabular}{l|c}
    \toprule
    Method & Avg. \\
    \midrule
    Task Arithmetic 
    & 68.00\zerogain \\
    \quad w/ GPTQ
    & 65.03\downgain{3.0} \\
    \quad w/o Anchor ($\alpha=0$)
    & 5.37\downgain{62.6} \\
    \rowcolor{FeatCalRow}\quad w/ \method~($\alpha=0.1$)
    & \textbf{74.09}\upgain{6.1} \\
    \midrule
    TIES-Merging 
    & 72.20\zerogain \\
    \quad w/ GPTQ
    & 65.03\downgain{7.2} \\
    \quad w/o Anchor ($\alpha=0$)
    & 4.57\downgain{67.6} \\
    \rowcolor{FeatCalRow}\quad w/ \method~($\alpha=0.01$)
    & \textbf{74.75}\upgain{2.5} \\
    \bottomrule
    \end{tabular}
    }
    \caption{
    Anchor ablation.
    }
    \label{tab:anchor_ablation}
    \vspace{-1.0em}
\end{wraptable}
}

\newcommand{\figBitwidthAnalysis}{%
\begin{wrapfigure}[12]{r}{0.60\columnwidth}
\centering
\vspace{-0.8em}
\includegraphics[width=\linewidth]{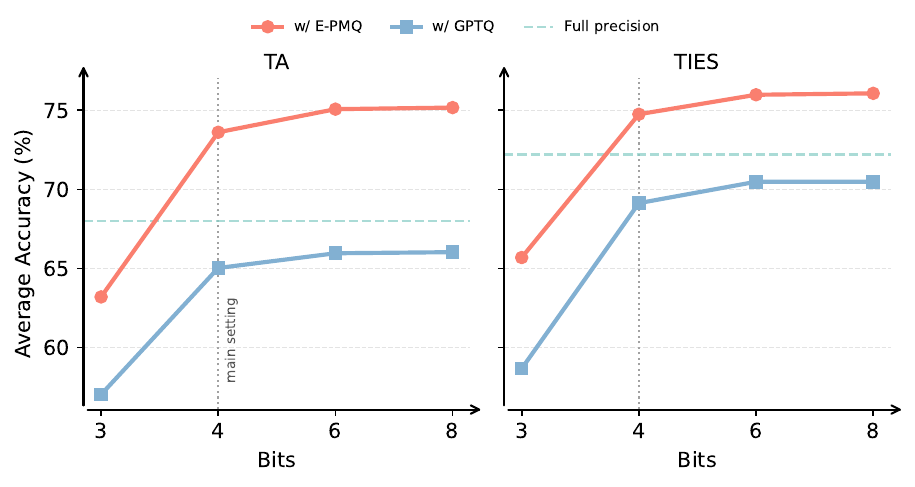}
\vspace{-2em}
\caption{
\small Bit-width analysis on CLIP-ViT-B/32.
E-PMQ consistently outperforms GPTQ from 3-bit to 8-bit.
}
\label{fig:bitwidth_analysis}
\end{wrapfigure}
}

\newcommand{\tabCalibBudgetTime}{%
\begin{table*}[h]
\centering
\caption{\small Calibration budget and quantization time.}
\label{tab:calib_budget_time}
\vspace{0.2em}
\setlength{\tabcolsep}{3.0pt}
\renewcommand{\arraystretch}{0.95}
\small
\resizebox{0.95\textwidth}{!}{
\begin{tabular}{lcccc@{\hspace{1.0em}}cccc@{\hspace{1.0em}}cccc}
\toprule
Method
& Samp. & Avg. & Time & Ratio
& Samp. & Avg. & Time & Ratio
& Samp. & Avg. & Time & Ratio \\
\midrule
Task Arithmetic
& -- & 68.00 & -- & --
& -- & 68.00 & -- & --
& -- & 68.00 & -- & -- \\
\midrule
\quad w/ GPTQ
& 64  & 64.86 & 29.2s  & 1.00$\times$
& 128 & 64.77 & 33.9s  & 1.00$\times$
& 256 & 65.03 & 71.0s  & 1.00$\times$ \\
\rowcolor{FeatCalRow}\quad w/ \method
& 64  & \textbf{72.23} & 52.8s   & 1.81$\times$
& 128 & \textbf{73.58} & 101.4s  & 2.99$\times$
& 256 & \textbf{73.61} & 172.2s  & 2.43$\times$ \\
\bottomrule
\end{tabular}
}
\vspace{-0.5em}
\end{table*}
}

\newcommand{\tabClipVitLFourteenPmqResults}{%
\begin{table}[h]
\centering
\caption{
Performance of 4-bit post-merge quantized CLIP-ViT-L/14 models on eight image-classification tasks.
All numbers are top-1 accuracy (\%). Gray arrows show changes over the corresponding full-precision merged checkpoints.
}
\label{tab:clip-vit-l14-pmq-results}
\setlength{\tabcolsep}{3.2pt}
\resizebox{\textwidth}{!}{
\begin{tabular}{l|cccccccc|c}
\toprule
Method & SUN397 & Cars & RESISC45 & EuroSAT & SVHN & GTSRB & MNIST & DTD & Avg. \\
\midrule
Simple Averaging 
& 72.5\zerogain & 81.5\zerogain & 82.2\zerogain & 90.0\zerogain & 81.6\zerogain & 74.0\zerogain & 96.6\zerogain & 61.8\zerogain & 80.0\zerogain \\

\quad w/ RTN 
& 71.6\downgain{0.9} & 80.4\downgain{1.1} & 81.2\downgain{1.0} & 83.2\downgain{6.8} & 75.6\downgain{6.0} & 68.8\downgain{5.2} & 96.6\zerogain & 61.5\downgain{0.3} & 77.4\downgain{2.6} \\

\quad w/ GPTQ 
& 72.0\downgain{0.5} & 80.5\downgain{1.0} & 81.8\downgain{0.4} & 86.5\downgain{3.5} & 80.8\downgain{0.8} & 69.0\downgain{5.0} & 96.7\upgain{0.1} & 61.7\downgain{0.1} & 78.6\downgain{1.4} \\

\quad w/ AWQ 
& 66.6\downgain{5.9} & 63.4\downgain{18.1} & 68.5\downgain{13.7} & 66.0\downgain{24.0} & 58.9\downgain{22.7} & 52.3\downgain{21.7} & 89.8\downgain{6.8} & 55.0\downgain{6.8} & 65.1\downgain{14.9} \\

\rowcolor{FeatCalRow}\quad w/ \method 
& 76.2\upgain{3.7} & 88.3\upgain{6.8} & 89.0\upgain{6.8} & 87.4\downgain{2.6} & 97.1\upgain{15.5} & 76.8\upgain{2.8} & 99.2\upgain{2.6} & 74.2\upgain{12.4} & \textbf{86.0}\upgain{6.0} \\

\midrule
Task Arithmetic 
& 72.0\zerogain & 79.0\zerogain & 80.5\zerogain & 86.0\zerogain & 87.5\zerogain & 83.5\zerogain & 98.0\zerogain & 58.8\zerogain & 80.7\zerogain \\

\quad w/ RTN 
& 71.0\downgain{1.0} & 77.0\downgain{2.0} & 79.6\downgain{0.9} & 77.1\downgain{8.9} & 84.6\downgain{3.0} & 72.7\downgain{10.8} & 97.9\downgain{0.1} & 58.8\zerogain & 77.3\downgain{3.4} \\

\quad w/ GPTQ 
& 71.3\downgain{0.8} & 77.2\downgain{1.8} & 80.1\downgain{0.4} & 79.7\downgain{6.3} & 87.2\downgain{0.3} & 73.2\downgain{10.3} & 98.1\upgain{0.1} & 59.5\upgain{0.7} & 78.3\downgain{2.4} \\

\quad w/ AWQ 
& 62.6\downgain{9.4} & 57.2\downgain{21.8} & 66.9\downgain{13.6} & 67.6\downgain{18.4} & 72.8\downgain{14.7} & 58.8\downgain{24.7} & 95.9\downgain{2.1} & 53.7\downgain{5.1} & 66.9\downgain{13.8} \\

\rowcolor{FeatCalRow}\quad w/ \method 
& 75.9\upgain{3.9} & 87.9\upgain{8.9} & 89.4\upgain{8.9} & 87.3\upgain{1.3} & 97.0\upgain{9.5} & 76.5\downgain{7.0} & 99.1\upgain{1.1} & 74.4\upgain{15.6} & \textbf{85.9}\upgain{5.2} \\

\midrule
TIES-Merging
& 74.7\zerogain & 83.3\zerogain & 86.4\zerogain & 91.3\zerogain & 89.7\zerogain & 85.2\zerogain & 97.8\zerogain & 63.9\zerogain & 84.0\zerogain \\

\quad w/ RTN
& 73.8\downgain{0.9} & 81.5\downgain{1.8} & 85.4\downgain{1.0} & 85.2\downgain{6.1} & 86.6\downgain{3.1} & 76.3\downgain{8.9} & 97.8\zerogain & 62.7\downgain{1.2} & 81.2\downgain{2.8} \\

\quad w/ GPTQ 
& 74.0\downgain{0.7} & 82.0\downgain{1.3} & 85.9\downgain{0.5} & 85.6\downgain{5.7} & 89.5\downgain{0.2} & 75.9\downgain{9.3} & 97.6\downgain{0.2} & 64.0\upgain{0.1} & 81.8\downgain{2.2} \\

\quad w/ AWQ 
& 70.4\downgain{4.4} & 69.2\downgain{14.2} & 77.3\downgain{9.1} & 75.0\downgain{16.3} & 77.2\downgain{12.5} & 63.5\downgain{21.7} & 95.4\downgain{2.4} & 59.4\downgain{4.5} & 73.4\downgain{10.6} \\

\rowcolor{FeatCalRow}\quad w/ \method 
& 76.0\upgain{1.3} & 87.9\upgain{4.6} & 89.1\upgain{2.7} & 87.8\downgain{3.5} & 97.0\upgain{7.3} & 76.4\downgain{8.8} & 99.1\upgain{1.3} & 74.8\upgain{10.9} & \textbf{86.0}\upgain{2.0} \\

\midrule
WUDI-Merging 
& 80.3\zerogain & 90.8\zerogain & 94.1\zerogain & 98.4\zerogain & 97.0\zerogain & 98.0\zerogain & 99.3\zerogain & 80.4\zerogain & \textbf{92.3}\zerogain \\

\quad w/ RTN
& 78.9\downgain{1.4} & 90.2\downgain{0.7} & 93.6\downgain{0.5} & 97.5\downgain{0.9} & 96.6\downgain{0.4} & 88.2\downgain{9.8} & 99.3\zerogain & 77.8\downgain{2.6} & 90.3\downgain{2.0} \\

\quad w/ GPTQ 
& 79.1\downgain{1.2} & 90.2\downgain{0.6} & 94.0\downgain{0.2} & 97.7\downgain{0.7} & 96.9\downgain{0.1} & 88.4\downgain{9.6} & 99.3\zerogain & 78.5\downgain{1.9} & 90.5\downgain{1.8} \\

\quad w/ AWQ 
& 74.1\downgain{6.2} & 78.5\downgain{12.3} & 85.4\downgain{8.7} & 92.6\downgain{5.8} & 94.9\downgain{2.1} & 84.3\downgain{13.7} & 98.4\downgain{0.9} & 71.6\downgain{8.8} & 85.0\downgain{7.3} \\

\rowcolor{FeatCalRow}\quad w/ \method 
& 79.9\downgain{0.4} & 91.0\upgain{0.2} & 95.1\upgain{1.0} & 97.82\downgain{0.6} & 97.3\upgain{0.3} & 88.9\downgain{9.1} & 99.3\zerogain & 80.8\upgain{0.4} & 91.3\downgain{1.0} \\
\bottomrule
\end{tabular}
}
\end{table}
}

\newcommand{\tabClipBThirtyTwoFourteenTaskFullPmq}{%
\begin{table*}[h]
\centering
\caption{
Full 14-task CLIP-ViT-B/32 results under 4-bit PMQ.
All numbers are top-1 accuracy (\%). Gray arrows in Avg. show changes over the corresponding full-precision merged checkpoints.
}
\label{tab:clip-b32-14task-full-pmq}
\setlength{\tabcolsep}{2.2pt}
\scriptsize
\resizebox{\textwidth}{!}{
\begin{tabular}{l|cccccccccccccc|c}
\toprule
Method & SUN397 & Cars & RESISC45 & EuroSAT & SVHN & GTSRB & MNIST & DTD & Flowers & PCAM & FER & Pets & STL10 & C100 & Avg. \\
\midrule
Simple Averaging
& 64.80 & 60.40 & 67.10 & 67.00 & 50.70 & 45.60 & 76.60 & 46.90 & 67.40 & 65.20 & 51.60 & 84.20 & 97.20 & 70.40 & 65.40\zerogain \\
\quad w/ RTN
& 61.63 & 53.75 & 62.06 & 53.07 & 50.48 & 42.64 & 73.41 & 43.51 & 60.56 & 52.38 & 39.91 & 86.40 & 96.45 & 64.97 & 60.09\downgain{5.3} \\
\quad w/ GPTQ
& 63.89 & 58.34 & 66.13 & 60.04 & 49.58 & 44.17 & 75.09 & 47.61 & 64.35 & 64.39 & 38.02 & 88.01 & 96.63 & 69.19 & 63.24\downgain{2.2} \\
\quad w/ AWQ
& 61.32 & 51.00 & 60.59 & 49.22 & 51.15 & 43.95 & 73.86 & 46.22 & 59.88 & 53.30 & 41.21 & 86.40 & 96.05 & 64.40 & 59.90\downgain{5.5} \\
\rowcolor{FeatCalRow}\quad w/ \method
& 65.65 & 63.95 & 75.63 & 64.00 & 91.46 & 54.32 & 97.89 & 57.39 & 74.03 & 75.44 & 39.51 & 90.30 & 95.10 & 74.09 & \textbf{72.77}\upgain{7.4} \\
\midrule
Task Arithmetic
& 41.80 & 33.20 & 47.30 & 55.40 & 46.50 & 48.40 & 88.70 & 37.00 & 38.60 & 64.10 & 46.10 & 65.90 & 84.60 & 41.70 & 52.80\zerogain \\
\quad w/ RTN
& 37.93 & 24.90 & 41.73 & 47.96 & 45.45 & 41.37 & 86.23 & 34.63 & 32.38 & 54.20 & 31.32 & 66.88 & 82.76 & 38.82 & 47.61\downgain{5.2} \\
\quad w/ GPTQ
& 40.34 & 31.05 & 45.76 & 51.44 & 46.44 & 44.19 & 87.43 & 36.70 & 35.92 & 58.14 & 28.71 & 68.60 & 83.58 & 40.50 & 49.92\downgain{2.9} \\
\quad w/ AWQ
& 37.97 & 27.16 & 42.56 & 47.04 & 44.58 & 40.91 & 87.26 & 34.89 & 34.62 & 57.72 & 31.53 & 67.87 & 83.55 & 38.50 & 48.30\downgain{4.5} \\
\rowcolor{FeatCalRow}\quad w/ \method
& 64.23 & 58.48 & 70.75 & 57.56 & 91.28 & 51.50 & 97.55 & 55.27 & 69.69 & 73.64 & 38.42 & 88.39 & 94.25 & 69.68 & \textbf{70.05}\upgain{17.3} \\
\midrule
TIES-Merging
& 62.20 & 54.60 & 65.30 & 63.00 & 65.70 & 63.90 & 92.60 & 49.90 & 58.20 & 77.10 & 54.90 & 81.40 & 94.80 & 62.40 & 67.60\zerogain \\
\quad w/ RTN
& 58.21 & 45.94 & 61.76 & 54.37 & 62.85 & 57.24 & 90.95 & 45.90 & 50.54 & 52.49 & 35.66 & 82.77 & 93.88 & 58.41 & 60.78\downgain{6.8} \\
\quad w/ GPTQ
& 60.51 & 53.17 & 64.70 & 58.15 & 65.62 & 57.85 & 92.29 & 49.31 & 55.60 & 59.66 & 31.36 & 84.11 & 94.45 & 61.60 & 63.46\downgain{4.1} \\
\quad w/ AWQ
& 58.42 & 48.09 & 61.27 & 52.52 & 64.59 & 56.48 & 91.10 & 46.17 & 50.20 & 51.56 & 37.73 & 82.56 & 93.45 & 57.01 & 60.80\downgain{6.8} \\
\rowcolor{FeatCalRow}\quad w/ \method
& 65.28 & 63.26 & 75.78 & 61.30 & 91.17 & 53.36 & 97.70 & 57.34 & 72.37 & 74.39 & 38.80 & 90.19 & 94.59 & 73.34 & \textbf{72.06}\upgain{4.5} \\
\midrule
WUDI-Merging
& 65.70 & 64.80 & 77.00 & 89.10 & 91.30 & 91.50 & 99.00 & 60.70 & 63.80 & 85.50 & 64.20 & 86.20 & 96.10 & 66.60 & \textbf{78.70}\zerogain \\
\quad w/ RTN
& 60.72 & 57.80 & 70.16 & 81.63 & 90.26 & 76.03 & 98.76 & 53.46 & 54.92 & 82.48 & 38.31 & 81.90 & 94.70 & 60.00 & 71.51\downgain{7.2} \\
\quad w/ GPTQ
& 63.77 & 62.77 & 77.52 & 85.19 & 91.05 & 77.57 & 99.03 & 58.14 & 61.26 & 85.06 & 37.52 & 85.09 & 95.66 & 65.33 & 74.64\downgain{4.1} \\
\quad w/ AWQ
& 61.75 & 60.09 & 70.76 & 79.37 & 90.45 & 77.09 & 98.84 & 55.59 & 55.68 & 80.97 & 37.81 & 81.06 & 94.95 & 61.67 & 71.86\downgain{6.8} \\
\rowcolor{FeatCalRow}\quad w/ \method
& 67.44 & 65.63 & 82.83 & 87.56 & 92.43 & 76.18 & 98.97 & 63.46 & 70.37 & 86.57 & 39.24 & 88.93 & 96.88 & 73.63 & 77.86\downgain{0.8} \\
\bottomrule
\end{tabular}
}
\end{table*}
}

\newcommand{\tabClipLFourteenFourteenTaskFullPmq}{%
\begin{table*}[h]
\centering
\caption{
Full 14-task CLIP-ViT-L/14 results under 4-bit PMQ.
All numbers are top-1 accuracy (\%). Gray arrows in Avg. show changes over the corresponding full-precision merged checkpoints.
}
\label{tab:clip-l14-14task-full-pmq}
\setlength{\tabcolsep}{2.2pt}
\scriptsize
\resizebox{\textwidth}{!}{
\begin{tabular}{l|cccccccccccccc|c}
\toprule
Method & SUN397 & Cars & RESISC45 & EuroSAT & SVHN & GTSRB & MNIST & DTD & Flowers & PCAM & FER & Pets & STL10 & C100 & Avg. \\
\midrule
Simple Averaging
& 71.20 & 79.00 & 78.70 & 80.40 & 71.30 & 64.60 & 94.30 & 58.70 & 81.90 & 74.20 & 54.80 & 94.60 & 99.30 & 82.40 & 77.50\zerogain \\
\quad w/ RTN
& 70.33 & 77.95 & 78.03 & 73.85 & 66.28 & 62.80 & 94.33 & 58.40 & 80.27 & 65.03 & 33.91 & 94.55 & 99.13 & 81.24 & 74.01\downgain{3.5} \\
\quad w/ GPTQ
& 70.67 & 77.74 & 77.89 & 76.33 & 70.44 & 62.28 & 94.44 & 58.83 & 80.71 & 75.09 & 37.85 & 94.88 & 99.26 & 81.86 & 75.59\downgain{1.9} \\
\quad w/ AWQ
& 66.16 & 62.32 & 64.25 & 52.59 & 44.29 & 46.40 & 83.53 & 53.35 & 74.01 & 57.20 & 35.87 & 92.45 & 96.71 & 68.24 & 64.10\downgain{13.4} \\
\rowcolor{FeatCalRow}\quad w/ \method
& 72.95 & 84.48 & 84.11 & 78.11 & 96.11 & 71.78 & 94.25 & 67.34 & 94.68 & 84.05 & 41.42 & 95.42 & 98.68 & 85.82 & \textbf{82.09}\upgain{4.6} \\
\midrule
Task Arithmetic
& 60.60 & 53.20 & 48.10 & 53.00 & 50.10 & 54.20 & 93.00 & 41.60 & 59.60 & 75.80 & 53.90 & 89.30 & 94.20 & 57.20 & 63.10\zerogain \\
\quad w/ RTN
& 58.54 & 49.79 & 46.97 & 45.52 & 44.14 & 46.40 & 92.08 & 41.91 & 57.68 & 50.02 & 39.97 & 88.44 & 93.51 & 54.60 & 57.83\downgain{5.3} \\
\quad w/ GPTQ
& 59.88 & 52.37 & 48.90 & 46.52 & 49.55 & 47.44 & 92.84 & 42.45 & 58.64 & 50.03 & 40.26 & 89.02 & 94.06 & 57.10 & 59.22\downgain{3.9} \\
\quad w/ AWQ
& 46.45 & 27.92 & 36.05 & 35.89 & 31.87 & 35.82 & 83.58 & 35.96 & 50.07 & 50.02 & 38.66 & 79.89 & 88.51 & 46.02 & 49.05\downgain{14.1} \\
\rowcolor{FeatCalRow}\quad w/ \method
& 72.75 & 84.85 & 84.75 & 80.52 & 95.54 & 71.87 & 98.63 & 66.76 & 93.71 & 78.61 & 40.42 & 95.48 & 98.61 & 84.86 & \textbf{81.95}\upgain{18.9} \\
\midrule
TIES-Merging
& 72.00 & 75.60 & 76.50 & 69.70 & 77.20 & 75.10 & 96.60 & 57.80 & 79.60 & 78.20 & 59.90 & 94.70 & 98.40 & 77.70 & 77.80\zerogain \\
\quad w/ RTN
& 71.03 & 73.22 & 74.95 & 61.07 & 73.39 & 66.94 & 96.26 & 57.29 & 78.29 & 56.34 & 40.50 & 94.14 & 98.30 & 75.97 & 72.69\downgain{5.1} \\
\quad w/ GPTQ
& 71.46 & 74.88 & 76.56 & 61.81 & 76.48 & 67.70 & 96.58 & 58.35 & 78.39 & 64.26 & 40.61 & 93.89 & 98.44 & 76.92 & 74.02\downgain{3.8} \\
\quad w/ AWQ
& 67.71 & 60.10 & 68.84 & 47.48 & 62.32 & 57.12 & 95.49 & 54.26 & 76.55 & 57.49 & 41.21 & 92.53 & 97.21 & 71.99 & 67.88\downgain{9.9} \\
\rowcolor{FeatCalRow}\quad w/ \method
& 73.06 & 84.96 & 84.95 & 80.96 & 95.81 & 72.22 & 98.74 & 67.82 & 93.84 & 80.98 & 41.50 & 95.72 & 98.65 & 85.84 & \textbf{82.50}\upgain{4.7} \\
\midrule
WUDI-Merging
& 76.70 & 87.60 & 90.40 & 95.40 & 94.80 & 95.70 & 99.20 & 71.40 & 95.40 & 86.70 & 70.70 & 96.20 & 99.10 & 84.50 & \textbf{88.80}\zerogain \\
\quad w/ RTN
& 75.71 & 86.36 & 89.51 & 92.63 & 93.93 & 85.24 & 99.13 & 70.11 & 94.08 & 84.28 & 40.89 & 95.75 & 98.93 & 83.02 & 84.97\downgain{3.8} \\
\quad w/ GPTQ
& 76.11 & 86.25 & 90.10 & 94.19 & 94.56 & 85.84 & 99.16 & 71.65 & 94.91 & 85.01 & 41.01 & 95.69 & 99.03 & 83.42 & 85.49\downgain{3.3} \\
\quad w/ AWQ
& 72.64 & 76.05 & 84.43 & 84.78 & 91.49 & 80.10 & 98.53 & 66.01 & 91.54 & 78.54 & 40.26 & 94.55 & 98.19 & 79.45 & 81.18\downgain{7.6} \\
\rowcolor{FeatCalRow}\quad w/ \method
& 77.80 & 89.27 & 92.70 & 95.56 & 95.77 & 87.05 & 99.19 & 75.27 & 96.31 & 81.14 & 41.39 & 96.02 & 99.23 & 86.45 & 86.65\downgain{2.2} \\
\bottomrule
\end{tabular}
}
\end{table*}
}

\newcommand{\tabClipBThirtyTwoTwentyTaskFullPmq}{%
\begin{table*}[h]
\centering
\caption{
Full 20-task CLIP-ViT-B/32 results under 4-bit PMQ.
All numbers are top-1 accuracy (\%). Gray arrows in Avg. show changes over the corresponding full-precision merged checkpoints.
}
\label{tab:clip-b32-20task-full-pmq}
\vspace{-0.5em}
\setlength{\tabcolsep}{1.0pt}
\renewcommand{\arraystretch}{0.78}
\tiny
\resizebox{\textwidth}{!}{
\begin{tabular}{l|cccccccccccccccccccc|c}
\toprule
Method 
& SUN & Cars & RES & Euro & SVHN & GTSRB & MNIST & DTD 
& Flwr & PCAM & FER & Pets & STL & C100 
& C10 & Food & FMNIST & EMNIST & KMNIST & SST2 & Avg. \\
\midrule
Simple Averaging
& 64.20 & 59.60 & 64.80 & 60.90 & 47.30 & 43.10 & 71.80 & 46.40 & 66.50 & 63.90 & 50.20 & 84.10 & 97.00 & 69.80 & 92.70 & 80.40 & 71.30 & 15.00 & 11.50 & 61.80 & 61.10\zerogain \\
\quad w/ RTN
& 60.95 & 53.12 & 59.24 & 46.26 & 48.30 & 40.59 & 68.96 & 42.29 & 59.47 & 51.07 & 39.82 & 85.83 & 96.35 & 64.01 & 90.03 & 76.90 & 71.32 & 16.15 & 13.05 & 56.89 & 57.03\downgain{4.1} \\
\quad w/ GPTQ
& 62.79 & 57.07 & 64.35 & 54.11 & 47.35 & 42.76 & 71.31 & 46.91 & 64.30 & 57.65 & 38.09 & 88.01 & 96.53 & 69.19 & 91.49 & 80.22 & 68.72 & 16.36 & 12.76 & 58.59 & 59.43\downgain{1.7} \\
\quad w/ AWQ
& 60.63 & 50.58 & 57.68 & 43.70 & 48.11 & 42.42 & 69.36 & 43.94 & 59.13 & 53.55 & 40.29 & 85.61 & 95.94 & 64.01 & 89.62 & 76.10 & 65.16 & 15.21 & 13.52 & 56.84 & 56.57\downgain{4.5} \\
\rowcolor{FeatCalRow}\quad w/ \method
& 64.59 & 60.63 & 73.95 & 60.52 & 87.93 & 53.44 & 93.78 & 56.33 & 72.26 & 73.57 & 40.39 & 90.22 & 95.25 & 72.42 & 94.58 & 82.45 & 85.80 & 28.36 & 28.17 & 65.57 & \textbf{69.01}\upgain{7.9} \\
\midrule
Task Arithmetic
& 20.40 & 12.20 & 25.60 & 25.60 & 30.90 & 29.80 & 78.00 & 22.30 & 21.10 & 53.20 & 34.30 & 42.40 & 71.00 & 29.50 & 64.10 & 15.10 & 67.00 & 17.00 & 15.40 & 51.20 & 36.30\zerogain \\
\quad w/ RTN
& 17.68 & 8.57 & 21.59 & 21.07 & 29.36 & 27.51 & 74.59 & 20.96 & 16.69 & 49.10 & 26.85 & 43.34 & 67.96 & 27.32 & 56.98 & 14.34 & 58.20 & 15.59 & 12.37 & 52.22 & 33.11\downgain{3.2} \\
\quad w/ GPTQ
& 19.83 & 11.65 & 24.52 & 23.52 & 30.83 & 27.19 & 77.07 & 21.60 & 19.99 & 49.62 & 23.04 & 46.61 & 70.21 & 29.30 & 60.88 & 15.09 & 63.87 & 16.56 & 12.75 & 55.02 & 34.96\downgain{1.3} \\
\quad w/ AWQ
& 17.72 & 9.66 & 22.87 & 23.07 & 29.54 & 27.58 & 76.21 & 21.44 & 18.47 & 52.62 & 24.92 & 43.80 & 69.26 & 27.41 & 59.15 & 13.29 & 61.93 & 15.59 & 12.81 & 52.28 & 33.98\downgain{2.3} \\
\rowcolor{FeatCalRow}\quad w/ \method
& 61.83 & 52.39 & 65.43 & 50.00 & 86.61 & 46.66 & 91.45 & 51.44 & 64.77 & 66.98 & 37.92 & 86.45 & 93.44 & 66.36 & 92.00 & 73.55 & 81.74 & 26.01 & 23.37 & 64.74 & \textbf{64.16}\upgain{27.9} \\
\midrule
TIES-Merging
& 51.00 & 36.20 & 47.80 & 45.10 & 58.20 & 57.70 & 92.10 & 40.60 & 44.80 & 66.90 & 47.30 & 73.10 & 89.90 & 51.30 & 86.30 & 50.10 & 76.50 & 21.00 & 19.70 & 55.90 & 55.60\zerogain \\
\quad w/ RTN
& 48.24 & 30.15 & 45.37 & 36.85 & 55.68 & 47.81 & 90.96 & 36.65 & 40.14 & 50.61 & 32.82 & 74.90 & 89.53 & 49.23 & 84.02 & 46.90 & 71.65 & 19.85 & 16.00 & 53.43 & 51.04\downgain{4.6} \\
\quad w/ GPTQ
& 49.96 & 35.22 & 47.62 & 42.19 & 58.39 & 50.53 & 91.88 & 40.48 & 43.03 & 57.45 & 25.23 & 76.42 & 89.93 & 50.63 & 84.62 & 48.66 & 75.89 & 20.30 & 17.02 & 56.18 & 53.08\downgain{2.5} \\
\quad w/ AWQ
& 47.50 & 31.75 & 45.48 & 42.89 & 57.11 & 49.74 & 91.72 & 38.09 & 38.88 & 52.19 & 29.83 & 74.19 & 89.35 & 47.98 & 83.04 & 44.11 & 73.65 & 20.39 & 16.73 & 53.82 & 51.42\downgain{4.2} \\
\rowcolor{FeatCalRow}\quad w/ \method
& 64.04 & 58.85 & 72.57 & 57.89 & 87.70 & 51.37 & 93.12 & 54.52 & 70.45 & 73.08 & 39.51 & 89.13 & 95.10 & 71.29 & 94.20 & 79.97 & 84.77 & 27.12 & 25.94 & 65.79 & \textbf{67.82}\upgain{12.2} \\
\midrule
WUDI-Merging
& 55.10 & 44.80 & 59.30 & 78.50 & 79.70 & 82.90 & 98.10 & 50.30 & 49.30 & 82.00 & 58.50 & 77.70 & 93.40 & 59.80 & 90.30 & 53.10 & 83.90 & 35.40 & 40.00 & 69.00 & 67.10\zerogain \\
\quad w/ RTN
& 49.86 & 37.50 & 53.76 & 69.70 & 78.45 & 69.52 & 97.57 & 44.04 & 40.62 & 72.16 & 36.67 & 75.44 & 92.20 & 55.83 & 87.85 & 45.74 & 84.61 & 31.79 & 21.17 & 67.49 & 60.60\downgain{6.5} \\
\quad w/ GPTQ
& 52.84 & 42.31 & 58.89 & 75.11 & 78.65 & 71.84 & 98.02 & 47.82 & 47.29 & 77.44 & 36.26 & 79.04 & 92.75 & 58.51 & 88.79 & 49.52 & 84.50 & 33.87 & 23.93 & 67.76 & 63.26\downgain{3.8} \\
\quad w/ AWQ
& 51.75 & 37.56 & 53.19 & 67.26 & 79.73 & 69.96 & 97.83 & 45.11 & 43.50 & 67.75 & 36.28 & 75.25 & 92.01 & 56.01 & 88.28 & 45.89 & 84.88 & 31.59 & 22.06 & 67.00 & 60.64\downgain{6.5} \\
\rowcolor{FeatCalRow}\quad w/ \method
& 64.37 & 57.41 & 79.08 & 85.44 & 89.25 & 73.90 & 98.02 & 56.91 & 61.13 & 85.90 & 38.98 & 87.24 & 96.30 & 70.85 & 94.37 & 72.08 & 87.22 & 36.78 & 29.32 & 69.52 & \textbf{71.70}\upgain{4.6} \\
\bottomrule
\end{tabular}
}
\vspace{-0.8em}
\end{table*}
}

\newcommand{\tabClipLFourteenTwentyTaskFullPmq}{%
\begin{table*}[t]
\centering
\caption{
Full 20-task CLIP-ViT-L/14 results under 4-bit PMQ.
All numbers are top-1 accuracy (\%). Gray arrows in Avg. show changes over the corresponding full-precision merged checkpoints.
}
\label{tab:clip-l14-20task-full-pmq}
\vspace{-0.5em}
\setlength{\tabcolsep}{1.0pt}
\renewcommand{\arraystretch}{0.78}
\tiny
\resizebox{\textwidth}{!}{
\begin{tabular}{l|cccccccccccccccccccc|c}
\toprule
Method 
& SUN & Cars & RES & Euro & SVHN & GTSRB & MNIST & DTD 
& Flwr & PCAM & FER & Pets & STL & C100 
& C10 & Food & FMNIST & EMNIST & KMNIST & SST2 & Avg. \\
\midrule
Simple Averaging
& 70.70 & 77.70 & 76.40 & 75.30 & 69.50 & 62.10 & 93.70 & 57.70 & 80.80 & 73.60 & 52.70 & 94.20 & 99.20 & 81.70 & 97.00 & 90.70 & 77.40 & 16.10 & 10.40 & 66.10 & 71.10\zerogain \\
\quad w/ RTN
& 69.90 & 76.84 & 76.10 & 69.74 & 65.49 & 60.88 & 93.61 & 57.07 & 79.10 & 59.23 & 33.51 & 94.14 & 99.04 & 80.82 & 96.83 & 91.82 & 74.09 & 15.61 & 9.22 & 65.02 & 68.40\downgain{2.7} \\
\quad w/ GPTQ
& 70.32 & 76.59 & 76.46 & 71.33 & 68.28 & 60.59 & 93.14 & 58.03 & 79.90 & 72.97 & 35.51 & 94.66 & 99.16 & 81.31 & 96.88 & 92.18 & 74.46 & 16.16 & 10.55 & 68.81 & 69.86\downgain{1.2} \\
\quad w/ AWQ
& 66.07 & 63.61 & 62.30 & 47.15 & 48.40 & 46.14 & 86.97 & 52.71 & 73.28 & 54.83 & 36.15 & 92.31 & 96.98 & 70.23 & 93.34 & 83.94 & 63.21 & 12.17 & 10.84 & 65.07 & 61.29\downgain{9.8} \\
\rowcolor{FeatCalRow}\quad w/ \method
& 72.12 & 82.74 & 83.48 & 79.67 & 94.50 & 71.14 & 97.93 & 65.16 & 92.31 & 80.59 & 40.90 & 95.31 & 98.83 & 84.11 & 98.16 & 92.68 & 90.27 & 49.53 & 13.02 & 70.40 & \textbf{77.64}\upgain{6.5} \\
\midrule
Task Arithmetic
& 17.70 & 17.40 & 45.90 & 89.90 & 59.50 & 59.30 & 96.10 & 36.60 & 41.30 & 81.70 & 49.10 & 65.70 & 81.80 & 28.00 & 76.50 & 24.30 & 80.50 & 62.10 & 73.30 & 57.70 & 57.20\zerogain \\
\quad w/ RTN
& 21.65 & 11.91 & 17.94 & 25.00 & 18.75 & 23.87 & 76.87 & 22.45 & 21.17 & 50.02 & 27.79 & 55.82 & 73.36 & 23.19 & 61.26 & 16.08 & 53.85 & 12.46 & 10.50 & 50.19 & 33.71\downgain{23.5} \\
\quad w/ GPTQ
& 22.64 & 13.72 & 18.81 & 26.30 & 19.21 & 24.79 & 78.34 & 22.45 & 21.56 & 50.02 & 29.51 & 57.26 & 75.83 & 25.32 & 64.81 & 16.34 & 56.88 & 12.07 & 10.63 & 50.36 & 34.84\downgain{22.4} \\
\quad w/ AWQ
& 12.23 & 5.43 & 13.08 & 24.30 & 13.29 & 19.83 & 63.04 & 18.88 & 15.76 & 50.02 & 26.33 & 38.81 & 66.45 & 16.15 & 48.13 & 10.49 & 39.91 & 10.78 & 10.21 & 50.14 & 27.66\downgain{29.5} \\
\rowcolor{FeatCalRow}\quad w/ \method
& 72.06 & 83.15 & 82.75 & 76.63 & 93.50 & 70.18 & 97.80 & 64.57 & 91.97 & 73.44 & 40.47 & 95.31 & 98.71 & 82.76 & 98.01 & 91.91 & 89.91 & 48.57 & 12.44 & 70.02 & \textbf{76.71}\upgain{19.5} \\
\midrule
TIES-Merging
& 64.40 & 56.60 & 49.10 & 42.10 & 67.20 & 56.80 & 95.30 & 46.20 & 64.90 & 78.30 & 54.90 & 91.30 & 95.70 & 62.90 & 92.90 & 70.50 & 82.00 & 19.90 & 10.90 & 56.90 & 63.00\zerogain \\
\quad w/ RTN
& 63.03 & 53.95 & 48.51 & 40.48 & 63.77 & 51.18 & 94.76 & 46.81 & 63.07 & 50.03 & 38.14 & 90.98 & 95.30 & 61.95 & 92.00 & 69.22 & 81.65 & 17.34 & 10.01 & 73.31 & 60.27\downgain{2.7} \\
\quad w/ GPTQ
& 63.89 & 55.34 & 48.97 & 40.19 & 66.72 & 51.81 & 95.36 & 47.87 & 65.56 & 50.09 & 38.56 & 90.76 & 95.60 & 62.88 & 92.56 & 70.72 & 81.56 & 17.94 & 9.88 & 75.34 & 61.08\downgain{1.9} \\
\quad w/ AWQ
& 60.02 & 44.10 & 44.32 & 33.59 & 60.25 & 45.90 & 95.88 & 45.37 & 64.42 & 50.08 & 39.37 & 87.95 & 94.46 & 60.35 & 90.73 & 64.43 & 77.84 & 19.09 & 11.07 & 71.61 & 58.04\downgain{5.0} \\
\rowcolor{FeatCalRow}\quad w/ \method
& 71.99 & 82.89 & 84.06 & 80.00 & 94.51 & 71.10 & 97.90 & 65.59 & 92.36 & 78.20 & 40.47 & 95.39 & 98.73 & 83.88 & 98.19 & 92.56 & 90.35 & 47.65 & 13.13 & 70.07 & \textbf{77.45}\upgain{14.5} \\
\midrule
WUDI-Merging
& 70.30 & 72.10 & 73.30 & 69.70 & 81.90 & 84.30 & 98.10 & 56.80 & 85.90 & 83.70 & 64.20 & 94.50 & 97.50 & 72.90 & 95.70 & 83.50 & 89.60 & 33.20 & 33.30 & 74.80 & 75.80\zerogain \\
\quad w/ RTN
& 69.19 & 68.75 & 71.27 & 62.07 & 80.09 & 72.41 & 98.05 & 55.59 & 83.67 & 74.26 & 38.92 & 94.17 & 97.25 & 71.49 & 95.32 & 80.92 & 89.64 & 30.03 & 13.35 & 78.36 & 71.24\downgain{4.6} \\
\quad w/ GPTQ
& 69.77 & 70.43 & 73.49 & 63.78 & 81.69 & 73.64 & 98.11 & 57.55 & 84.88 & 76.66 & 39.20 & 94.03 & 97.29 & 72.45 & 95.53 & 82.37 & 89.98 & 30.90 & 13.39 & 78.64 & 72.19\downgain{3.6} \\
\quad w/ AWQ
& 66.49 & 57.02 & 68.78 & 55.52 & 78.56 & 66.70 & 98.19 & 56.12 & 81.64 & 74.29 & 39.36 & 92.61 & 96.78 & 67.36 & 94.22 & 77.79 & 86.65 & 30.93 & 14.12 & 78.09 & 69.06\downgain{6.7} \\
\rowcolor{FeatCalRow}\quad w/ \method
& 76.61 & 84.74 & 90.65 & 94.48 & 93.96 & 84.35 & 98.73 & 70.74 & 94.99 & 82.87 & 40.44 & 95.80 & 99.20 & 84.86 & 98.18 & 91.17 & 91.98 & 48.37 & 17.02 & 79.63 & \textbf{80.94}\upgain{5.1} \\
\bottomrule
\end{tabular}
}
\vspace{-0.8em}
\end{table*}
}

\newcommand{\tabAlphaAblationFull}{%
\begin{table*}[t]
\centering
\caption{
Anchor-strength ablation on CLIP-ViT-B/32 8-task setting under 4-bit PMQ.
All numbers are top-1 accuracy (\%). Gray arrows in Avg. show changes over the corresponding full-precision merged checkpoints.
}
\label{tab:alpha_ablation_full}
\setlength{\tabcolsep}{2.8pt}
\renewcommand{\arraystretch}{0.92}
\resizebox{\textwidth}{!}{
\begin{tabular}{ll|cccccccc|c}
\toprule
Method & $\alpha$ 
& SUN397 & Cars & RESISC45 & EuroSAT & SVHN & GTSRB & MNIST & DTD & Avg. \\
\midrule
Task Arithmetic 
& --
& 57.10 & 55.70 & 64.90 & 76.70 & 77.90 & 68.50 & 96.10 & 47.20 & 68.00\zerogain \\

\quad w/ GPTQ
& --
& 55.60 & 53.54 & 63.83 & 69.11 & 77.87 & 56.99 & 95.88 & 47.45 & 65.03\downgain{3.0} \\

\multirow{5}{*}{\quad w/ \method}
& 0
& 0.29 & 0.51 & 2.17 & 16.41 & 8.72 & 2.39 & 9.78 & 2.71 & 5.37\downgain{62.6} \\

& 0.01
& 67.03 & 64.36 & 78.52 & 66.30 & 94.75 & 57.03 & 98.92 & 61.97 & 73.61\upgain{5.6} \\

& \cellcolor{FeatCalRow}0.1
& \cellcolor{FeatCalRow}67.09 
& \cellcolor{FeatCalRow}64.03 
& \cellcolor{FeatCalRow}79.24 
& \cellcolor{FeatCalRow}68.33 
& \cellcolor{FeatCalRow}94.55 
& \cellcolor{FeatCalRow}58.81 
& \cellcolor{FeatCalRow}98.72 
& \cellcolor{FeatCalRow}61.91 
& \cellcolor{FeatCalRow}\textbf{74.09}\upgain{6.1} \\

& 1
& 66.12 & 63.52 & 78.37 & 70.78 & 92.71 & 61.27 & 98.31 & 60.48 & 73.94\upgain{5.9} \\

& 10
& 64.60 & 60.66 & 76.37 & 76.37 & 89.92 & 64.37 & 97.59 & 55.80 & 73.21\upgain{5.2} \\

\midrule
TIES-Merging
& --
& 67.10 & 64.20 & 74.10 & 76.80 & 77.70 & 69.40 & 94.10 & 54.00 & 72.20\zerogain \\

\quad w/ GPTQ
& --
& 55.60 & 53.54 & 63.83 & 69.11 & 77.87 & 56.99 & 95.88 & 47.45 & 65.03\downgain{7.2} \\

\multirow{5}{*}{\quad w/ \method}
& 0
& 0.29 & 0.52 & 2.79 & 9.81 & 8.90 & 3.28 & 9.74 & 1.22 & 4.57\downgain{67.6} \\

& \cellcolor{FeatCalRow}0.01
& \cellcolor{FeatCalRow}67.61 
& \cellcolor{FeatCalRow}66.65 
& \cellcolor{FeatCalRow}80.46 
& \cellcolor{FeatCalRow}67.19 
& \cellcolor{FeatCalRow}94.66 
& \cellcolor{FeatCalRow}59.20 
& \cellcolor{FeatCalRow}98.96 
& \cellcolor{FeatCalRow}63.24 
& \cellcolor{FeatCalRow}\textbf{74.75}\upgain{2.6} \\

& 0.1
& 67.47 & 66.40 & 80.05 & 67.67 & 93.98 & 61.31 & 98.65 & 62.13 & 74.71\upgain{2.5} \\

& 1
& 66.72 & 65.02 & 78.24 & 69.33 & 91.71 & 60.71 & 97.71 & 60.32 & 73.72\upgain{1.5} \\

& 10
& 65.14 & 60.75 & 75.78 & 71.04 & 86.89 & 63.25 & 96.25 & 55.80 & 71.86\downgain{0.3} \\
\bottomrule
\end{tabular}
}
\end{table*}
}

\begin{abstract}
Low-resource deployment constraints have made model quantization essential for
deploying neural networks while preserving performance. Meanwhile, model
merging has become an increasingly practical low-resource strategy for
integrating multiple task- or domain-specialized experts into a single
model without joint training or multi-model serving. Together, quantization and model merging enable an efficient low-resource
deployment pipeline by integrating multiple experts into one low-bit model. We formulate this setting as \emph{Post-Merge Quantization} (PMQ).
We show that directly applying post-training quantization (PTQ) to a merged
model is unreliable because two distinct deviations are coupled: the
\emph{quantization deviation} introduced by low-bit reconstruction and the
\emph{expert-relative merging deviation} inherited from model merging. To
mitigate these deviations, we propose \emph{E-PMQ}, an expert-guided PMQ
framework that uses source expert weights to provide expert-guided output
targets during layer-wise calibration, together with merged-weight anchoring to
stabilize the calibration and preserve the integrated behavior of the merged
model. On CLIP-ViT-B/32 eight-task merging, E-PMQ improves 4-bit GPTQ from
65.0\% to 73.6\% under Task Arithmetic and from 69.1\% to 74.8\% under
TIES-Merging. On harder settings, E-PMQ improves GPTQ from 34.8\% to 76.7\% on
20-task CLIP-ViT-L/14 and from 78.26\% to 83.34\% on FLAN-T5-base GLUE. These
results demonstrate that E-PMQ enables effective post-merge quantization and
low-bit deployment.

\end{abstract}
\vspace{-1.58em}

\section{Introduction}

Low-resource deployment constraints have made model quantization essential for
deploying neural networks while preserving performance. Low-bit post-training
quantization (PTQ) is one of the most practical techniques for this setting, as
it converts full-precision weights into low-bit representations using only a
small calibration set and without expensive end-to-end retraining. Existing PTQ
methods have achieved strong results for independently trained models,
where the full-precision model is typically treated as a reliable reconstruction
target during layer-wise quantization~\citep{gptq,awq,smoothquant,adaround,brecq}.

Model merging is also an increasingly practical low-resource strategy. Instead
of jointly training a multi-task model or serving multiple experts,
merging integrates several task- or domain-specialized models into a
single model~\citep{weight_averaging,task_arithmetic,fisher_merging,ties_merging,dare,cheng2025wudi}.
This makes merging attractive for resource-constrained adaptation and
deployment: the resulting model can combine capabilities from multiple
experts while avoiding multi-model serving. However, a merged model is not
necessarily an independently optimized multi-task model. Since it is obtained
through parameter composition, it may already deviate from the expert behaviors
that merging aims to preserve.

These two low-resource techniques naturally meet in deployment: after experts
are merged into a single model, the resulting model may still need to be
quantized for low-bit inference. We formulate this setting as
\emph{Post-Merge Quantization} (PMQ), where the quantization target is a merged
model rather than an independently trained model. This distinction is
important because naive PMQ couples two distinct deviations. The first is the
\emph{quantization deviation} introduced by low-bit reconstruction. The second
is the \emph{expert-relative merging deviation} inherited from model merging.
Directly applying ordinary PTQ methods such as GPTQ~\citep{gptq} to a merged
model only reconstructs the merged model itself, and therefore treats this
potentially deviated model as the sole target. As a result, naive PMQ may
preserve expert-relative merging deviations and further compound them with
quantization deviation, making the standard merge-then-quantize pipeline
unreliable, especially under aggressive low-bit settings.

To mitigate these deviations, we propose \emph{E-PMQ}, an expert-guided PMQ
framework with merged-weight anchoring. During layer-wise calibration, E-PMQ
uses source expert weights to provide expert-guided output targets. These
targets introduce expert-relative guidance into the quantization process, rather
than passively reconstructing only the merged model. Together with this
expert guidance, merged-weight anchoring stabilizes the calibration and
preserves the integrated behavior of the merged model. The expert
models are accessed only during the post-merge calibration stage. After
quantization, the deployed model remains a single low-bit merged model,
without experts or additional inference-time modules. Figure~\ref{fig:main}
illustrates this distinction.

\figMain

Experiments show that E-PMQ consistently improves low-bit merged models across
vision and text settings. On CLIP-ViT-B/32 eight-task merging, E-PMQ improves 4-bit GPTQ from
65.0\% to 73.6\% under Task Arithmetic and from 69.1\% to 74.8\% under
TIES-Merging. The gains remain strong in harder settings: under Task
Arithmetic, E-PMQ improves GPTQ from 34.8\% to 76.7\% on 20-task
CLIP-ViT-L/14 and from 78.26\% to 83.34\% on FLAN-T5-base GLUE. Further
experiments show consistent gains across merging methods, task scales,
modalities, and quantization bit-widths.

We summarize the main contributions of this work as follows:

\par\noindent\ding{182}\enspace
We formulate \emph{Post-Merge Quantization} (PMQ) as a distinct low-bit
deployment setting for merged models, and identify a key failure mode of
naive PMQ: directly reconstructing the merged model couples the
\emph{quantization deviation} introduced by low-bit reconstruction with the
\emph{expert-relative merging deviation} inherited from model merging.

\par\noindent\ding{183}\enspace
We introduce \emph{E-PMQ}, an expert-guided PMQ framework that uses source
expert weights to provide expert-guided output targets during layer-wise
calibration, together with merged-weight anchoring to stabilize the calibration
and preserve the integrated behavior of the merged model.

\par\noindent\ding{184}\enspace
We validate E-PMQ on CLIP and FLAN-T5, showing consistent gains over naive PMQ
baselines such as GPTQ across merging methods, task scales, modalities, and
quantization bit-widths.

\section{Related Work}

\paragraph{Model Merging.}
Model merging composes multiple specialized models into a single model without joint training or deploying one model per task. Existing methods include weight averaging, Fisher merging, task arithmetic, TIES-Merging, DARE, and adaptive or data-free task-vector approaches~\citep{weight_averaging,fisher_merging,task_arithmetic,ties_merging,dare}. Recent surveys and systems work frame model fusion as a scalable alternative to repeatedly training or serving many experts~\citep{zhou2026model,zhou2025democratizing,wang2026mergepipe,wang2025model}, while broader fusion methods explore preference- or distillation-based composition~\citep{gu2025infifpo,wang2025infigfusion}. These works focus on building, scaling, or managing merged models; our work instead studies how to quantize an already merged model more reliably.

\paragraph{Post-training quantization.}
Post-training quantization compresses a trained full-precision model into low-bit weights without end-to-end retraining, typically through calibration-based rounding, scaling, or layer-wise reconstruction~\citep{adaround,brecq,gptq,awq,smoothquant,zeroquant}. Ordinary PTQ generally assumes that the full-precision model is a reliable target to preserve, which is natural for independently trained models but less reliable for merged models. Low-precision training and inference recipes further highlight the importance of numerical efficiency for scalable deployment~\citep{wang2025infir2comprehensivefp8training}. Our work studies PMQ, where naive merge-then-quantize baselines apply ordinary PTQ such as GPTQ to a merged model. Instead of only reconstructing the merged model, E-PMQ uses source expert weights during layer-wise quantization to construct expert-guided calibration targets and anchors the solution to the merged model for stability.

\section{Preliminaries and Problem Formulation}
\label{sec:preliminaries}

\paragraph{Notation.}
Let $\{W_i\}_{i=1}^{K}$ denote $K$ task-specialized expert
models, and $W_m = \mathcal{M}(\{W_i\}_{i=1}^{K})$ be the merged
model produced by a merging algorithm $\mathcal{M}$. We use $W_i^\ell$,
$W_m^\ell$, and $Q^\ell$ to denote the layer-$\ell$ weights of expert $i$, the
merged model, and the quantized model, respectively. Let
$\mathcal{D}_{\mathrm{cal}}$ be a small calibration set, and let
$X^\ell \in \mathbb{R}^{d_{\mathrm{in}}\times n}$ denote the calibration
activations entering layer $\ell$, where $n$ is the number of calibration
tokens. The feasible set of $b$-bit quantized weights is denoted by
$\mathcal{Q}_b$.

\paragraph{Post-Training Quantization.}
Post-training quantization compresses a full-precision model into a low-bit
model using a small calibration set, without end-to-end retraining. For a
generic full-precision model $W$, a PTQ algorithm produces a quantized model
\begin{equation}
    Q
    =
    \mathcal{A}_{\mathrm{ptq}}(W; \mathcal{D}_{\mathrm{cal}}),
    \label{eq:ptq_definition}
\end{equation}
where $\mathcal{A}_{\mathrm{ptq}}$ denotes the PTQ algorithm and $Q$ is the
resulting $b$-bit model.

Following the layer-wise reconstruction formulation used in
GPTQ~\citep{gptq}, a reconstruction-based PTQ method minimizes the following
layer-wise objective:
\begin{equation}
    \min_{Q^\ell \in \mathcal{Q}_b}
    \left\|
        Q^\ell X^\ell - W^\ell X^\ell
    \right\|_F^2 .
    \label{eq:ptq_objective}
\end{equation}
Accordingly, we characterize the layer-wise \textbf{quantization deviation} as
\begin{equation}
    \Delta_{\mathrm{quant}}^\ell(X^\ell)
    =
    Q^\ell X^\ell - W^\ell X^\ell .
    \label{eq:quant_deviation_general}
\end{equation}

\paragraph{Model Merging.}
Model merging combines multiple task- or domain-specialized experts into
a single model without joint training or deploying one model per task:
\begin{equation}
    W_m
    =
    \mathcal{M}\left(\{W_i\}_{i=1}^{K}\right).
    \label{eq:model_merge}
\end{equation}
Since $W_m$ is obtained by parameter composition, its intermediate
representations may deviate from those of the original experts. Prior work has
observed such representation-level discrepancy between merged models and
source experts during model merging~\citep{yang2024representation}. Following
this view, we characterize the layer-wise \textbf{expert-relative merging
deviation} in the output space. We use $X^\ell$ as a common layer-wise
input, which isolates the output discrepancy induced by
different layer weights under the same inputs. The deviation of the
merged layer from expert $i$ is
\begin{equation}
    \Delta_{\mathrm{merge},i}^{\ell}(X^\ell)
    =
    W_m^\ell X^\ell - W_i^\ell X^\ell .
    \label{eq:merge_deviation}
\end{equation}
This term measures how far the merged model has moved away from the
behavior of each source expert before quantization is applied.

\paragraph{Post-Merge Quantization.}
In this work, we formulate post-merge quantization, where the goal is to obtain
a low-bit model after merging. PMQ produces a quantized merged model
\begin{equation}
    Q_m
    =
    \mathcal{A}_{\mathrm{pmq}}(W_m, \{W_i\}_{i=1}^{K};
    \mathcal{D}_{\mathrm{cal}}),
    \label{eq:pmq_definition}
\end{equation}
where $\mathcal{A}_{\mathrm{pmq}}$ denotes a post-merge quantization algorithm.
A straightforward solution is to directly apply a standard PTQ algorithm to the
merged model:
\begin{equation}
    Q_m^{\mathrm{naive}}
    =
    \mathcal{A}_{\mathrm{ptq}}(W_m; \mathcal{D}_{\mathrm{cal}}).
    \label{eq:naive_pmq}
\end{equation}
At layer $\ell$, following the GPTQ-style reconstruction objective, naive PMQ
minimizes
\begin{equation}
    \min_{Q^\ell \in \mathcal{Q}_b}
    \left\|
       Q^\ell X^\ell - W_m^\ell X^\ell
    \right\|_F^2 .
    \label{eq:naive_pmq_objective}
\end{equation}
However, this objective treats the full-precision merged model as a
reliable standalone reconstruction target. This assumption is problematic in
PMQ because the merged model may already contain expert-relative merging
deviations before quantization.

To make this deviation explicit, consider the output deviation of the
quantized merged layer with respect to expert $i$:
\begin{equation}
    Q^\ell X^\ell - W_i^\ell X^\ell
    =
    \underbrace{
    Q^\ell X^\ell - W_m^\ell X^\ell
    }_{\text{quantization deviation}}
    +
    \underbrace{
    W_m^\ell X^\ell - W_i^\ell X^\ell
    }_{\text{expert-relative merging deviation}} .
    \label{eq:pmq_deviation_decomposition}
\end{equation}
The first term is introduced by low-bit quantization and corresponds to the
standard reconstruction deviation considered by PTQ methods. The second term is
inherited from model merging: it measures how the full-precision merged layer
deviates from each source expert and is therefore invisible to naive PMQ
objectives that only reconstruct $W_m^\ell$. This distinction makes PMQ
fundamentally different from quantizing an independently trained model. In PMQ,
the quantized model should not merely approximate the merged model; it
must also avoid further compounding the expert-relative deviations that already
exist after merging. Otherwise, the quantization deviation is added on top of
the merging deviation, and their accumulated effect can perturb intermediate
representations as they propagate through the network, ultimately degrading
downstream task performance. This observation motivates a PMQ method that goes
beyond passive reconstruction of the merged model and explicitly uses
source experts to guide the quantization of the merged model.

\section{Method}
\label{sec:method}

\subsection{Overview}

We propose \emph{E-PMQ}, an expert-guided post-merge quantization framework. 
Given a full-precision merged model and its source experts, E-PMQ performs layer-wise quantization in forward order. 
When quantizing layer $\ell$, earlier layers have already been quantized or fixed, so the calibration activations reflect the activation distribution encountered by the current partially quantized merged model.

For layer $\ell$, E-PMQ uses the merged weight $W_m^\ell$, expert weights $\{W_i^\ell\}_{i=1}^{K}$, and calibration activations $\{X_i^\ell\}_{i=1}^{K}$. 
Here, $X_i^\ell$ denotes the layer-wise calibration activation collected from the current quantization trajectory using the calibration subset associated with expert $i$.

E-PMQ uses expert weights to construct expert-guided output targets on these calibration activations, while anchoring the quantized weight to the full-precision merged weight for stability.

\subsection{Expert-Guided Objective}

Following GPTQ-style reconstruction-based PTQ, E-PMQ formulates layer-wise
quantization as an output reconstruction problem on calibration activations.
To mitigate expert-relative merging deviation during quantization, we use the
corresponding source expert weight to construct the layer-wise output target:
\begin{equation}
    Y_i^\ell
    =
    W_i^\ell X_i^\ell .
    \label{eq:expert_guided_target}
\end{equation}
This gives the expert-guided reconstruction objective:
\begin{equation}
    \min_{Q^\ell \in \mathcal{Q}_b}
    \sum_{i=1}^{K}
    \left\|
    Q^\ell X_i^\ell
    -
    W_i^\ell X_i^\ell
    \right\|_F^2 ,
    \label{eq:expert_guided_reconstruction}
\end{equation}
where $\mathcal{Q}_b$ denotes the $b$-bit quantization space. Unlike
standard merged-model reconstruction, which treats the full-precision merged
output as the target, this objective uses the source experts to provide
output targets for the quantized merged layer. Since the inputs $X_i^\ell$
are collected from the current quantization trajectory, the reconstruction is
performed on the activation distribution that the quantized merged model will
actually encounter.

However, expert-guided reconstruction alone may over-correct the merged layer,
especially when different experts contain partially conflicting task-specific
updates. To preserve the integrated behavior produced by model merging, we
add a merged-weight anchor:
\begin{equation}
    \min_{Q^\ell \in \mathcal{Q}_b}
    \sum_{i=1}^{K}
    \left\|
    Q^\ell X_i^\ell
    -
    W_i^\ell X_i^\ell
    \right\|_F^2
    +
    \lambda^\ell
    \left\|
    Q^\ell - W_m^\ell
    \right\|_F^2 .
    \label{eq:epmq_objective}
\end{equation}
The first term mitigates expert-relative merging deviation during quantization
by matching expert-induced output targets. The second term keeps the
quantized weight close to the full-precision merged weight, preventing the
solution from drifting toward isolated experts and helping preserve the
merged model's integrated behavior.

\subsection{Adaptive Merged-Weight Anchoring}

The anchor strength $\lambda^\ell$ controls the trade-off between expert-guided output targets and preservation of the merged model. 
Since different layers can have different activation scales, we use an activation-adaptive anchor:
\begin{equation}
    \lambda^\ell
    =
    \frac{\alpha}{d_\ell}
    \operatorname{Tr}
    \left(
    \sum_{i=1}^{K}
    X_i^\ell (X_i^\ell)^\top
    \right)
    =
    \frac{\alpha}{d_\ell}
    \sum_{i=1}^{K}
    \left\|
    X_i^\ell
    \right\|_F^2 ,
    \label{eq:adaptive_lambda}
\end{equation}
where $d_\ell$ is the input dimension of layer $\ell$ and $\alpha$ is a global scaling hyperparameter. 
This choice scales the anchor with the total calibration activation energy of the layer and adds diagonal loading to the corresponding quadratic form.

\subsection{GPTQ-Style Solver}
\label{sec:gptq_style_solver}

Eq.~\eqref{eq:epmq_objective} is constrained to the discrete low-bit space
$\mathcal{Q}_b$, so the deployed quantized weight cannot be obtained by simply
using a continuous closed-form solution. In practice, E-PMQ solves the
layer-wise objective with a GPTQ-style sequential rounding solver. The solver
keeps the implementation structure of GPTQ while using the expert-guided
objective and merged-weight anchoring defined above.

To expose the quadratic statistics used by the solver, define
\begin{equation}
    H_i^\ell
    =
    X_i^\ell (X_i^\ell)^\top,
    \qquad
    H_q^\ell
    =
    \sum_{i=1}^{K} H_i^\ell .
    \label{eq:epmq_feature_stats}
\end{equation}
Under the E-PMQ objective, the corresponding effective curvature and
right-hand side are
\begin{equation}
    H_{\mathrm{E\text{-}PMQ}}^\ell
    =
    H_q^\ell + \lambda^\ell I,
    \qquad
    R_{\mathrm{E\text{-}PMQ}}^\ell
    =
    \sum_{i=1}^{K}
    W_i^\ell H_i^\ell
    +
    \lambda^\ell W_m^\ell .
    \label{eq:epmq_solver_statistics}
\end{equation}
The term $\sum_{i=1}^{K} W_i^\ell H_i^\ell$ is induced by the
expert-guided output targets, while $\lambda^\ell I$ and
$\lambda^\ell W_m^\ell$ come from merged-weight anchoring. The full procedure is summarized in 
Appendix~\ref{app:algorithm}. We provide the continuous relaxation, stationary
condition, and closed-form relaxed optimizer in
Appendix~\ref{app:continuous_relaxation}.

\section{Experiments}
\subsection{Experimental Setup}
\label{sec:exp-setup}

\paragraph{Benchmarks.}
We evaluate E-PMQ using the FusionBench model-merging benchmark suite~\citep{tang2025fusionbench}.
For vision experiments, we use CLIP-ViT-B/32 and CLIP-ViT-L/14~\citep{radford2021learning} on the standard 8-task image-classification suite, including SUN397, Stanford Cars, RESISC45, EuroSAT, SVHN, GTSRB, MNIST, and DTD~\citep{xiao2010sun,krause2013cars,cheng2017remote,helber2019eurosat,netzer2011reading,stallkamp2011german,lecun1998gradient,cimpoi2014describing}.
We further evaluate 14-task and 20-task CLIP suites to test scalability to more merged tasks.
For language experiments, we use FLAN-T5-base~\citep{raffel2020exploring,wei2022finetuned,chung2024scaling} on eight GLUE tasks~\citep{wang2018glue}: CoLA, MNLI, MRPC, QNLI, QQP, RTE, SST-2, and STS-B.
We report task scores and average performance across tasks.

\paragraph{Merging and quantization methods.}
For CLIP, we evaluate Simple Averaging, Task Arithmetic, TIES-Merging, and WUDI-Merging as upstream merging methods.
For FLAN-T5, we evaluate Task Arithmetic and TIES-Merging.
After obtaining the full-precision merged model, we compare E-PMQ with naive PMQ baselines, including RTN, GPTQ~\citep{gptq}, and AWQ~\citep{awq}.
These baselines correspond to naive PMQ pipelines that quantize the merged model directly.

\paragraph{Quantization protocol.}
Unless otherwise specified, all quantized models use 4-bit weight-only quantization.
The main experiments use 256 calibration samples per task; thus, a \(K\)-task merged model uses \(256K\) calibration samples in total.
E-PMQ performs layer-wise quantization in forward order and uses the same calibration data as the PTQ baselines.
Implementation details, including batch size, group size, anchor hyperparameters, and so on, are provided in Appendix~\ref{app:implementation_details}.

\subsection{Main CLIP Results}

\tabClipVitBThirtyTwoPmqResults

Table~\ref{tab:clip-vit-b32-pmq-results} presents the main 8-task CLIP-ViT-B/32 results.
Naive PMQ baselines often lose accuracy after 4-bit quantization, especially when the upstream merger is relatively weak.
E-PMQ gives the best average accuracy among quantized methods for Simple Averaging, Task Arithmetic, and TIES-Merging, improving over GPTQ by 11.4, 8.6, and 5.7 points, respectively.
For WUDI-Merging, the full-precision model is already substantially stronger, leaving less room for calibration; in this setting, E-PMQ remains close to the full-precision model and competitive with naive PMQ baselines. The main pattern is consistent with our PMQ motivation.
When the merged model is a weak reconstruction target, directly quantizing it will preserve expert-relative merging deviation and compound them with low-bit quantization deviation.
E-PMQ is most beneficial in these cases because source expert weights provide expert-guided output targets during calibration, while merged-weight anchoring prevents destructive over-correction. Full CLIP-ViT-L/14 8-task results are provided in Appendix~\ref{app:clip_l14_8task} as a backbone-scaling experiment; the same trend holds on the larger CLIP backbone, indicating that the gains are not specific to ViT-B/32.

\subsection{Extended CLIP Results}

\tabClipExtendedAvg

We next increase the number of merged tasks.
Table~\ref{tab:clip-extended-avg} summarizes results across the 8-task, 14-task, and 20-task CLIP settings on both CLIP-ViT-B/32 and CLIP-ViT-L/14.
These results suggest that PMQ becomes harder as the merger must absorb more experts.
With more tasks, the merged model is more likely to contain interference among expert updates, making direct reconstruction of the merged output less reliable.

For E-PMQ, the largest gains appear in the 20-task setting. Under Task Arithmetic, E-PMQ improves the average accuracy by more than 27 points over the full-precision merged model on CLIP-ViT-B/32 and by 19.5 points on CLIP-ViT-L/14. This indicates that E-PMQ is not merely compressing the merged model; through source expert guidance, it also corrects expert-relative deviations that are already present before quantization.
Full per-task results are provided in Appendix~\ref{app:clip_14task_full} and Appendix~\ref{app:clip_20task_full}.

\subsection{Results on FLAN-T5}

\tabFlanTFiveGlueResults
Table~\ref{tab:flan-t5-glue-results} reports the results on FLAN-T5 merged
models under 4-bit PMQ. This experiment evaluates whether E-PMQ generalizes
beyond CLIP-based vision models to language-model merging. Across both Task
Arithmetic and TIES-Merging, E-PMQ consistently outperforms RTN, GPTQ, and AWQ,
showing that its gains are not tied to a specific architecture, modality, or PTQ
baseline.

Under Task Arithmetic, RTN and GPTQ slightly degrade the average score of the
full-precision merged model, while E-PMQ improves it from 78.79 to 83.34. Under
TIES-Merging, E-PMQ further improves the average score from 79.98 to 83.48 and
achieves the best overall performance. These results suggest that
language-model merging also produces imperfect reconstruction targets for naive
PMQ. By using source-expert guidance and a merged-weight anchor, E-PMQ mitigates
both expert-relative merging deviation and quantization deviation, reducing
their accumulation under low-bit quantization.

\subsection{Results on LLM}

\tabLlamaPmqResults

Table~\ref{tab:llama-pmq-results} reports the results on Llama-3.1 models
merged by Task Arithmetic under 4-bit PMQ. This experiment further evaluates
whether E-PMQ remains effective on larger language models beyond FLAN-T5. We
evaluate two model scales, Llama-3.1-3B and Llama-3.1-8B, on a mixture of
mathematical reasoning, general reasoning, instruction-following, and
code-generation benchmarks. Additional implementation details for LLM
quantization are provided in Appendix~\ref{app:implementation_details_llm}.

Across both model scales, E-PMQ achieves the best average performance among all
quantized variants. On Llama-3.1-3B, E-PMQ improves the average score from
58.71 with GPTQ and 59.07 with AWQ to 60.27. On Llama-3.1-8B, E-PMQ improves
the average score from 61.66 with GPTQ and 62.27 with AWQ to 62.91. These
consistent gains show that E-PMQ is effective not only for CLIP-based vision
models and FLAN-T5, but also for larger language models. By using source-expert
guidance together with merged-weight anchoring, E-PMQ provides stronger
post-merge quantization than directly applying standard PTQ baselines to the
merged model.

\subsection{Anchor Ablation}

\tabAnchorAblation
\figAnchorAblation

Figure~\ref{fig:ablation_analysis} and Table~\ref{tab:anchor_ablation} isolate the role of merged-weight anchoring in \method{}.
Removing the anchor by setting \(\alpha=0\) causes severe collapse: the average accuracy drops from 68.00 to 5.37 under Task Arithmetic and from 72.20 to 4.57 under TIES-Merging.
This shows that expert-guided output targets alone do not define a stable low-bit quantization objective; without anchoring, the quantized solution can move too far from the merged model and lose the integrated behavior obtained by merging.

For positive anchor strengths, \method{} remains stable across a reasonable range of \(\alpha\) and consistently outperforms direct GPTQ, as shown in Figure~\ref{fig:alpha_sensitivity}.
The best value varies slightly across merging methods, but the trend is robust: merged-weight anchoring is a necessary regularizer rather than a minor tuning detail.
Additional per-task results under Task Arithmetic and TIES-Merging are provided in Appendix~\ref{app:alpha_ablation_full}.

\subsection{Bit-Width Analysis}

\figBitwidthAnalysis

Figure~\ref{fig:bitwidth_analysis} compares E-PMQ with GPTQ under different bit-widths. E-PMQ consistently outperforms GPTQ from 3-bit to 8-bit under both Task Arithmetic and TIES-Merging. The largest gains appear in the more aggressive low-bit regimes, where quantization deviation is more severe and naive reconstruction of the merged model is least reliable. This result suggests that E-PMQ remains useful across multiple low-bit deployment regimes.

\subsection{Calibration Budget and Quantization Cost}
We finally examine the trade-off between calibration budget, quantization-stage computation, and final 4-bit accuracy.
The calibration budget is measured as samples per task; in the 8-task setting, \(n\) samples per task correspond to \(8n\) total calibration images.
With only 64 samples per task, E-PMQ reaches 72.23 average accuracy, outperforming GPTQ with 256 samples per task by 7.20 points.

This result suggests that expert-guided targets provide a more informative calibration signal than merged-model reconstruction alone.
E-PMQ therefore trades additional pre-deployment computation for better use of limited calibration data and stronger low-bit merged-model quality.
This extra cost is incurred only during quantization. After quantization, E-PMQ has the same single-model inference form, parameter count, and bit-width as the corresponding GPTQ baseline.

\tabCalibBudgetTime

\section{Conclusion}

We studied \emph{Post-Merge Quantization} (PMQ), a low-bit deployment setting for merged models. PMQ differs from ordinary PTQ because the full-precision merged model may already contain deviations from the source experts. Directly applying ordinary PTQ to this model can therefore preserve an imperfect reconstruction target and compound expert-relative merging deviation with low-bit quantization deviation.

We proposed \emph{E-PMQ}, which constructs expert-guided calibration targets from source expert weights and uses merged-weight anchoring to stabilize low-bit calibration. Experiments on CLIP-based vision merging and FLAN-T5 language merging show consistent gains over naive PMQ baselines such as GPTQ across merging methods, task scales, and bit-widths. Further analyses confirm the necessity of merged-weight anchoring and show that E-PMQ improves low-bit merged-model quality while preserving the same single-model inference-time deployment form.

\section*{Acknowledgments}
This paper is fully supported by grants from the Research Grants Council of the Hong Kong Special Administrative Region, China (Project No. T41-517/25-N and 15228325 )

{\small
\bibliographystyle{plainnat}
\bibliography{references}

@article{zhou2026model,
  title={Model Fusion for Scalable and Sustainable Artificial Intelligence: A Review and Outlook},
  author={Zhou, Qi and Zhang, Yiming and Gu, Yanggan and Wang, Yuanyi and Yan, Zhaoyi and Li, Zhen and Chung, Chi Yung and Yang, Hongxia},
  journal={Journal of Modern Power Systems and Clean Energy},
  volume={14},
  number={1},
  pages={37--49},
  year={2026},
  publisher={SGEPRI}
}

@article{zhou2025democratizing,
  title={Democratizing AI through model fusion: A comprehensive review and future directions},
  author={Zhou, Qi and Zhang, Yiming and Gu, Yanggan and Wang, Yuanyi and Sang, Zhijie and Yan, Zhaoyi and Li, Zhen and Zhang, Shengyu and Wu, Fei and Yang, Hongxia},
  journal={Nexus},
  year={2025},
  publisher={Elsevier}
}

@misc{wang2026mergepipe,
      title={MergePipe: A Budget-Aware Parameter Management System for Scalable LLM Merging}, 
      author={Yuanyi Wang and Yanggan Gu and Zihao Wang and Kunxi Li and Yifan Yang and Zhaoyi Yan and Congkai Xie and Jianmin Wu and Hongxia Yang},
      year={2026},
      eprint={2602.13273},
      archivePrefix={arXiv},
      primaryClass={cs.DB},
      url={https://arxiv.org/abs/2602.13273}, 
}

@article{wang2025model,
  title={Model merging scaling laws in large language models},
  author={Wang, Yuanyi and Gu, Yanggan and Zhang, Yiming and Zhou, Qi and Yan, Zhaoyi and Xie, Congkai and Wang, Xinyao and Yuan, Jianbo and Yang, Hongxia},
  journal={arXiv preprint arXiv:2509.24244},
  year={2025}
}

@article{gu2025infifpo,
  title={InfiFPO: Implicit model fusion via preference optimization in large language models},
  author={Gu, Yanggan and Wang, Yuanyi and Yan, Zhaoyi and Zhang, Yiming and Zhou, Qi and Wu, Fei and Yang, Hongxia},
  journal={arXiv preprint arXiv:2505.13878},
  year={2025}
}

@article{wang2025infigfusion,
  title={Infigfusion: Graph-on-logits distillation via efficient gromov-wasserstein for model fusion},
  author={Wang, Yuanyi and Yan, Zhaoyi and Zhang, Yiming and Zhou, Qi and Gu, Yanggan and Wu, Fei and Yang, Hongxia},
  journal={arXiv preprint arXiv:2505.13893},
  year={2025}
}

@misc{wang2025infir2comprehensivefp8training,
      title={InfiR2: A Comprehensive FP8 Training Recipe for Reasoning-Enhanced Language Models}, 
      author={Wenjun Wang and Shuo Cai and Congkai Xie and Mingfa Feng and Yiming Zhang and Zhen Li and Kejing Yang and Ming Li and Jiannong Cao and Hongxia Yang},
      year={2025},
      eprint={2509.22536},
      archivePrefix={arXiv},
      primaryClass={cs.CL},
      url={https://arxiv.org/abs/2509.22536}, 
}

@article{tang2025fusionbench,
  title={FusionBench: A Unified Library and Comprehensive Benchmark for Deep Model Fusion},
  author={Tang, Anke and Shen, Li and Luo, Yong and Yang, Enneng and Hu, Han and Zhang, Lefei and Du, Bo and Tao, Dacheng},
  journal={Journal of Machine Learning Research},
  volume={26},
  number={307},
  pages={1--38},
  year={2025}
}

@inproceedings{radford2021learning,
  title={Learning Transferable Visual Models From Natural Language Supervision},
  author={Radford, Alec and Kim, Jong Wook and Hallacy, Chris and Ramesh, Aditya and Goh, Gabriel and Agarwal, Sandhini and Sastry, Girish and Askell, Amanda and Mishkin, Pamela and Clark, Jack and Krueger, Gretchen and Sutskever, Ilya},
  booktitle={International Conference on Machine Learning},
  pages={8748--8763},
  year={2021}
}

@article{raffel2020exploring,
  title={Exploring the Limits of Transfer Learning with a Unified Text-to-Text Transformer},
  author={Raffel, Colin and Shazeer, Noam and Roberts, Adam and Lee, Katherine and Narang, Sharan and Matena, Michael and Zhou, Yanqi and Li, Wei and Liu, Peter J.},
  journal={Journal of Machine Learning Research},
  volume={21},
  number={140},
  pages={1--67},
  year={2020}
}

@article{wei2022finetuned,
  title={Finetuned Language Models are Zero-Shot Learners},
  author={Wei, Jason and Bosma, Maarten and Zhao, Vincent Y. and Guu, Kelvin and Yu, Adams Wei and Lester, Brian and Du, Nan and Dai, Andrew M. and Le, Quoc V.},
  journal={International Conference on Learning Representations},
  year={2022}
}

@article{chung2024scaling,
  title={Scaling Instruction-Finetuned Language Models},
  author={Chung, Hyung Won and Hou, Le and Longpre, Shayne and Zoph, Barret and Tay, Yi and Fedus, William and Li, Eric and Wang, Xuezhi and Dehghani, Mostafa and Brahma, Siddhartha and Webson, Albert and Gu, Shixiang Shane and Dai, Zhuyun and Suzgun, Mirac and Chen, Xinyun and Chowdhery, Aakanksha and Narang, Sharan and Mishra, Gaurav and Yu, Adams Wei and Zhao, Vincent and Huang, Yanping and Dai, Andrew and Yu, Hongkun and Petrov, Slav and Chi, Ed H. and Dean, Jeff and Devlin, Jacob and Roberts, Adam and Zhou, Denny and Le, Quoc V. and Wei, Jason},
  journal={Journal of Machine Learning Research},
  volume={25},
  number={70},
  pages={1--53},
  year={2024}
}

@inproceedings{wang2018glue,
  title={GLUE: A Multi-Task Benchmark and Analysis Platform for Natural Language Understanding},
  author={Wang, Alex and Singh, Amanpreet and Michael, Julian and Hill, Felix and Levy, Omer and Bowman, Samuel R.},
  booktitle={International Conference on Learning Representations},
  year={2019}
}

@inproceedings{weight_averaging,
  title={Model soups: averaging weights of multiple fine-tuned models improves accuracy without increasing inference time},
  author={Wortsman, Mitchell and Ilharco, Gabriel and Gadre, Samir Yitzhak and Roelofs, Rebecca and Gontijo-Lopes, Raphael and Morcos, Ari S. and Namkoong, Hongseok and Farhadi, Ali and Carmon, Yair and Kornblith, Simon and Schmidt, Ludwig},
  booktitle={International Conference on Machine Learning},
  year={2022}
}

@inproceedings{fisher_merging,
  title={Merging Models with Fisher-Weighted Averaging},
  author={Matena, Michael S. and Raffel, Colin A.},
  booktitle={Advances in Neural Information Processing Systems},
  year={2022}
}

@inproceedings{task_arithmetic,
  title={Editing Models with Task Arithmetic},
  author={Ilharco, Gabriel and Ribeiro, Marco Tulio and Wortsman, Mitchell and Gururangan, Suchin and Schmidt, Ludwig and Hajishirzi, Hannaneh and Farhadi, Ali},
  booktitle={International Conference on Learning Representations},
  year={2023}
}

@inproceedings{ties_merging,
  title={TIES-Merging: Resolving Interference When Merging Models},
  author={Yadav, Prateek and Tam, Derek and Choshen, Leshem and Raffel, Colin and Bansal, Mohit},
  booktitle={Advances in Neural Information Processing Systems},
  year={2023}
}

@misc{dare,
      title={Language Models are Super Mario: Absorbing Abilities from Homologous Models as a Free Lunch}, 
      author={Le Yu and Bowen Yu and Haiyang Yu and Fei Huang and Yongbin Li},
      year={2024},
      eprint={2311.03099},
      archivePrefix={arXiv},
      primaryClass={cs.CL},
      url={https://arxiv.org/abs/2311.03099}, 
}

@inproceedings{adaround,
  title={Up or Down? Adaptive Rounding for Post-Training Quantization},
  author={Nagel, Markus and Amjad, Rana Ali and Van Baalen, Mart and Louizos, Christos and Blankevoort, Tijmen},
  booktitle={International Conference on Machine Learning},
  year={2020}
}

@inproceedings{brecq,
  title={BRECQ: Pushing the Limit of Post-Training Quantization by Block Reconstruction},
  author={Li, Yuhang and Gong, Ruihao and Tan, Xu and Yang, Yang and Hu, Peng and Zhang, Qi and Yu, Fengwei and Wang, Wei and Gu, Shi},
  booktitle={International Conference on Learning Representations},
  year={2021}
}

@inproceedings{gptq,
  title={GPTQ: Accurate Post-Training Quantization for Generative Pre-trained Transformers},
  author={Frantar, Elias and Ashkboos, Saleh and Hoefler, Torsten and Alistarh, Dan},
  booktitle={International Conference on Learning Representations},
  year={2023}
}

@inproceedings{awq,
  title={AWQ: Activation-aware Weight Quantization for LLM Compression and Acceleration},
  author={Lin, Ji and Tang, Jiaming and Tang, Haotian and Yang, Shang and Chen, Wei-Ming and Wang, Wei-Chen and Xiao, Guangxuan and Dang, Xingyu and Gan, Chuang and Han, Song},
  booktitle={Proceedings of Machine Learning and Systems},
  year={2024}
}

@inproceedings{smoothquant,
  title={SmoothQuant: Accurate and Efficient Post-Training Quantization for Large Language Models},
  author={Xiao, Guangxuan and Lin, Ji and Seznec, Mickael and Wu, Hao and Demouth, Julien and Han, Song},
  booktitle={International Conference on Machine Learning},
  year={2023}
}

@inproceedings{zeroquant,
  title={ZeroQuant: Efficient and Affordable Post-Training Quantization for Large-Scale Transformers},
  author={Yao, Zhewei and Aminabadi, Reza Yazdani and Zhang, Minjia and Wu, Xiaoxia and Li, Conglong and He, Yuxiong},
  booktitle={Advances in Neural Information Processing Systems},
  year={2022}
}

@inproceedings{xiao2010sun,
  title={SUN Database: Large-scale Scene Recognition from Abbey to Zoo},
  author={Xiao, Jianxiong and Hays, James and Ehinger, Krista A. and Oliva, Aude and Torralba, Antonio},
  booktitle={IEEE Conference on Computer Vision and Pattern Recognition},
  year={2010}
}

@inproceedings{krause2013cars,
  title={3D Object Representations for Fine-Grained Categorization},
  author={Krause, Jonathan and Stark, Michael and Deng, Jia and Fei-Fei, Li},
  booktitle={IEEE International Conference on Computer Vision Workshops},
  year={2013}
}

@inproceedings{cheng2017remote,
  title={Remote Sensing Image Scene Classification: Benchmark and State of the Art},
  author={Cheng, Gong and Han, Junwei and Lu, Xiaoqiang},
  booktitle={Proceedings of the IEEE},
  volume={105},
  number={10},
  pages={1865--1883},
  year={2017}
}

@article{helber2019eurosat,
  title={EuroSAT: A Novel Dataset and Deep Learning Benchmark for Land Use and Land Cover Classification},
  author={Helber, Patrick and Bischke, Benjamin and Dengel, Andreas and Borth, Damian},
  journal={IEEE Journal of Selected Topics in Applied Earth Observations and Remote Sensing},
  volume={12},
  number={7},
  pages={2217--2226},
  year={2019}
}

@inproceedings{netzer2011reading,
  title={Reading Digits in Natural Images with Unsupervised Feature Learning},
  author={Netzer, Yuval and Wang, Tao and Coates, Adam and Bissacco, Alessandro and Wu, Bo and Ng, Andrew Y.},
  booktitle={NeurIPS Workshop on Deep Learning and Unsupervised Feature Learning},
  year={2011}
}

@inproceedings{stallkamp2011german,
  title={The German Traffic Sign Recognition Benchmark: A multi-class classification competition},
  author={Stallkamp, Johannes and Schlipsing, Marc and Salmen, Jan and Igel, Christian},
  booktitle={International Joint Conference on Neural Networks},
  year={2011}
}

@article{lecun1998gradient,
  title={Gradient-Based Learning Applied to Document Recognition},
  author={LeCun, Yann and Bottou, L{\'e}on and Bengio, Yoshua and Haffner, Patrick},
  journal={Proceedings of the IEEE},
  volume={86},
  number={11},
  pages={2278--2324},
  year={1998}
}

@inproceedings{cimpoi2014describing,
  title={Describing Textures in the Wild},
  author={Cimpoi, Mircea and Maji, Subhransu and Kokkinos, Iasonas and Mohamed, Sammy and Vedaldi, Andrea},
  booktitle={IEEE Conference on Computer Vision and Pattern Recognition},
  year={2014}
}

@inproceedings{nilsback2008automated,
  title={Automated Flower Classification over a Large Number of Classes},
  author={Nilsback, Maria-Elena and Zisserman, Andrew},
  booktitle={Indian Conference on Computer Vision, Graphics and Image Processing},
  year={2008}
}

@inproceedings{veeling2018rotation,
  title={Rotation Equivariant CNNs for Digital Pathology},
  author={Veeling, Bastiaan S. and Linmans, Jasper and Winkens, Jim and Cohen, Taco and Welling, Max},
  booktitle={Medical Image Computing and Computer Assisted Intervention},
  year={2018}
}

@article{goodfellow2015challenges,
  title={Challenges in Representation Learning: A Report on Three Machine Learning Contests},
  author={Goodfellow, Ian J. and Erhan, Dumitru and Carrier, Pierre Luc and Courville, Aaron and Mirza, Mehdi and Hamner, Ben and Cukierski, Will and Tang, Yichuan and Thaler, David and Lee, Dong-Hyun and Zhou, Yingbo and Ramaiah, Chetan and Feng, Fangxiang and Li, Ruifan and Wang, Xiaojie and Athanasakis, Dimitris and Shawe-Taylor, John and Milakov, Maxim and Park, John and Ionescu, Radu and Popescu, Marius and Grozea, Cristian and Bergstra, James and Xie, Jingjing and Romaszko, Lukasz and Xu, Bing and Chuang, Zhang and Bengio, Yoshua},
  journal={Neural Networks},
  volume={64},
  pages={59--63},
  year={2015}
}

@inproceedings{parkhi2012cats,
  title={Cats and Dogs},
  author={Parkhi, Omkar M. and Vedaldi, Andrea and Zisserman, Andrew and Jawahar, C. V.},
  booktitle={IEEE Conference on Computer Vision and Pattern Recognition},
  year={2012}
}

@article{coates2011analysis,
  title={An Analysis of Single-Layer Networks in Unsupervised Feature Learning},
  author={Coates, Adam and Ng, Andrew Y. and Lee, Honglak},
  journal={International Conference on Artificial Intelligence and Statistics},
  year={2011}
}

@techreport{krizhevsky2009learning,
  title={Learning Multiple Layers of Features from Tiny Images},
  author={Krizhevsky, Alex},
  institution={University of Toronto},
  year={2009}
}

@inproceedings{bossard2014food,
  title={Food-101 -- Mining Discriminative Components with Random Forests},
  author={Bossard, Lukas and Guillaumin, Matthieu and Van Gool, Luc},
  booktitle={European Conference on Computer Vision},
  year={2014}
}

@article{xiao2017fashion,
  title={Fashion-MNIST: A Novel Image Dataset for Benchmarking Machine Learning Algorithms},
  author={Xiao, Han and Rasul, Kashif and Vollgraf, Roland},
  journal={arXiv preprint arXiv:1708.07747},
  year={2017}
}

@inproceedings{cohen2017emnist,
  title={EMNIST: Extending MNIST to handwritten letters},
  author={Cohen, Gregory and Afshar, Saeed and Tapson, Jonathan and van Schaik, Andre},
  booktitle={International Joint Conference on Neural Networks},
  year={2017}
}

@article{clanuwat2018deep,
  title={Deep Learning for Classical Japanese Literature},
  author={Clanuwat, Tarin and Bober-Irizar, Mikel and Kitamoto, Asanobu and Lamb, Alex and Yamamoto, Kazuaki and Ha, David},
  journal={arXiv preprint arXiv:1812.01718},
  year={2018}
}

@inproceedings{socher2013recursive,
  title={Recursive Deep Models for Semantic Compositionality Over a Sentiment Treebank},
  author={Socher, Richard and Perelygin, Alex and Wu, Jean and Chuang, Jason and Manning, Christopher D. and Ng, Andrew Y. and Potts, Christopher},
  booktitle={Conference on Empirical Methods in Natural Language Processing},
  year={2013}
}

@inproceedings{cheng2025wudi,
  title={Whoever Started the Interference Should End It: Guiding Data-Free Model Merging via Task Vectors},
  author={Cheng, Runxi and Xiong, Feng and Wei, Yongxian and Zhu, Wanyun and Yuan, Chun},
  booktitle={Proceedings of the 42nd International Conference on Machine Learning},
  series={Proceedings of Machine Learning Research},
  volume={267},
  pages={10121--10143},
  year={2025},
  publisher={PMLR}
}

@misc{yang2024representation,
      title={Representation Surgery for Multi-Task Model Merging}, 
      author={Enneng Yang and Li Shen and Zhenyi Wang and Guibing Guo and Xiaojun Chen and Xingwei Wang and Dacheng Tao},
      year={2024},
      eprint={2402.02705},
      archivePrefix={arXiv},
      primaryClass={cs.LG},
      url={https://arxiv.org/abs/2402.02705}, 
}
}

\newpage
\appendix

\section{Limitations}
\label{sec:limitations}

E-PMQ has several limitations. First, it requires access to the source experts during the pre-deployment quantization stage. If only the merged model is available, E-PMQ cannot construct expert-guided targets and reduces to direct post-merge quantization. This requirement also introduces additional quantization-stage memory and compute compared with naive PMQ baselines, since expert weights are used to form layer-wise output targets.

The cost of E-PMQ scales with the number of experts \(K\) and the calibration budget. This makes the method most suitable for scenarios where experts are available during compression and additional pre-deployment computation is acceptable. Importantly, this extra cost is not incurred at inference time: after quantization, E-PMQ produces a single low-bit merged model and does not require experts, calibration data, or additional inference-time modules.

E-PMQ also depends on the representativeness of the calibration set. If the calibration samples poorly cover the task distributions associated with the experts, the expert-guided targets may provide a weaker calibration signal. Finally, our experiments focus on CLIP and FLAN-T5. Extending E-PMQ to larger-scale LLMs, more diverse modalities, and more complex merging scenarios remains an important direction for future work.

\section{Algorithm}
\label{app:algorithm}

\begin{algorithm}[h]
\small
\caption{E-PMQ for Post-Merge Quantization}
\label{alg:epmq}
\begin{algorithmic}[1]
\REQUIRE Experts \(\{W_i\}_{i=1}^{K}\), merging operator \(\mathcal{M}\), calibration subsets \(\{\mathcal{D}_{\mathcal{cal},i}\}_{i=1}^{K}\), bit-width \(b\), anchor scale \(\alpha\).
\STATE Obtain \(W_m=\mathcal{M}(\{W_i\}_{i=1}^{K})\) and initialize the current model from \(W_m\).
\FOR{each layer \(\ell=1,\ldots,L\)}
    \FOR{each expert/task \(i=1,\ldots,K\)}
        \STATE Collect \(X_i^\ell\) by running the current partially quantized merged model on \(\mathcal{D}_{\mathcal{cal},i}\), and set \(H_i^\ell=X_i^\ell(X_i^\ell)^\top\).
    \ENDFOR
    \STATE Set \(H_q^\ell=\sum_{i=1}^{K}H_i^\ell\) and \(\lambda^\ell=\frac{\alpha}{d_\ell}\sum_{i=1}^{K}\|X_i^\ell\|_F^2\).
    \STATE Form \(H_{\mathrm{E\text{-}PMQ}}^\ell=H_q^\ell+\lambda^\ell I\) and \(R_{\mathrm{E\text{-}PMQ}}^\ell=\sum_{i=1}^{K}W_i^\ell H_i^\ell+\lambda^\ell W_m^\ell\).
    \STATE Use a GPTQ-style discrete rounding solver to obtain \(Q^\ell\in\mathcal{Q}_b\) under the E-PMQ objective.
    \STATE Replace layer \(\ell\) in the current model with \(Q^\ell\).
\ENDFOR
\RETURN Quantized merged model \(Q_m\).
\end{algorithmic}
\end{algorithm}

\section{Continuous Relaxation and Solver Statistics}
\label{app:continuous_relaxation}

We provide the continuous relaxation of the E-PMQ objective and derive the
quadratic statistics used by the GPTQ-style solver. Although the deployed
weight is obtained by sequential rounding in the discrete low-bit space, the
continuous relaxation clarifies how expert-guided output targets and
merged-weight anchoring modify the layer-wise quadratic problem.

Recall the E-PMQ objective for layer $\ell$:
\begin{equation}
    \min_{Q^\ell \in \mathcal{Q}_b}
    \sum_{i=1}^{K}
    \left\|
    Q^\ell X_i^\ell
    -
    W_i^\ell X_i^\ell
    \right\|_F^2
    +
    \lambda^\ell
    \left\|
    Q^\ell - W_m^\ell
    \right\|_F^2 .
    \label{eq:app_epmq_objective}
\end{equation}
To obtain a continuous relaxation, we ignore the discrete constraint
$Q^\ell \in \mathcal{Q}_b$. Define
\begin{equation}
    H_i^\ell
    =
    X_i^\ell (X_i^\ell)^\top,
    \qquad
    H_q^\ell
    =
    \sum_{i=1}^{K} H_i^\ell .
    \label{eq:app_epmq_feature_stats}
\end{equation}
Then the relaxed objective can be written as the following quadratic form:
\begin{equation}
    \sum_{i=1}^{K}
    \operatorname{Tr}
    \left[
    (Q^\ell - W_i^\ell)
    H_i^\ell
    (Q^\ell - W_i^\ell)^\top
    \right]
    +
    \lambda^\ell
    \left\|
    Q^\ell - W_m^\ell
    \right\|_F^2 .
    \label{eq:app_epmq_quadratic_form}
\end{equation}
Taking the derivative with respect to $Q^\ell$ and setting it to zero gives
the stationary condition
\begin{equation}
    Q^\ell
    \left(
    H_q^\ell + \lambda^\ell I
    \right)
    =
    \sum_{i=1}^{K}
    W_i^\ell H_i^\ell
    +
    \lambda^\ell W_m^\ell .
    \label{eq:app_epmq_stationary}
\end{equation}
Therefore, the continuous relaxed optimizer is
\begin{equation}
    Q_{\mathrm{cont}}^\ell
    =
    \left(
    \sum_{i=1}^{K}
    W_i^\ell H_i^\ell
    +
    \lambda^\ell W_m^\ell
    \right)
    \left(
    H_q^\ell + \lambda^\ell I
    \right)^{-1}.
    \label{eq:app_epmq_continuous_solution}
\end{equation}
When $\lambda^\ell > 0$, the anchoring term adds diagonal loading, making
$H_q^\ell + \lambda^\ell I$ positive definite and the inverse well defined.

The continuous solution in Eq.~\eqref{eq:app_epmq_continuous_solution} is not
the deployed quantized weight. In practice, E-PMQ uses the corresponding
quadratic form inside a GPTQ-style sequential rounding solver. The effective
curvature used by the solver is
\begin{equation}
    H_{\mathrm{E\text{-}PMQ}}^\ell
    =
    H_q^\ell + \lambda^\ell I,
    \label{eq:app_epmq_curvature}
\end{equation}
and the expert-guided right-hand side is
\begin{equation}
    R_{\mathrm{E\text{-}PMQ}}^\ell
    =
    \sum_{i=1}^{K}
    W_i^\ell H_i^\ell
    +
    \lambda^\ell W_m^\ell .
    \label{eq:app_epmq_rhs}
\end{equation}
Here, the term $\sum_{i=1}^{K} W_i^\ell H_i^\ell$ is induced by
expert-guided output targets, while $\lambda^\ell I$ and
$\lambda^\ell W_m^\ell$ arise from merged-weight anchoring. These statistics
are then used by the GPTQ-style sequential rounding procedure described in
Algorithm~\ref{alg:epmq}.

\section{Additional CLIP Results}
\label{app:additional_clip_results}

In addition to the 8-task CLIP-ViT-B/32 results in the main text, we report full per-task results for the larger backbone and extended task suites. 
The extended 14-task and 20-task settings include additional datasets such as Flowers102, PCAM, FER2013, Oxford-IIIT Pet, STL10, CIFAR100, CIFAR10, Food101, Fashion-MNIST, EMNIST Letters, KMNIST, and Rendered SST2~\citep{nilsback2008automated,veeling2018rotation,goodfellow2015challenges,parkhi2012cats,coates2011analysis,krizhevsky2009learning,bossard2014food,xiao2017fashion,cohen2017emnist,clanuwat2018deep,socher2013recursive,radford2021learning}.

\subsection{Full 8-Task Results on CLIP-ViT-L/14}
\label{app:clip_l14_8task}

\tabClipVitLFourteenPmqResults

Table~\ref{tab:clip-vit-l14-pmq-results} reports the full 8-task CLIP-ViT-L/14 results under 4-bit PMQ.
E-PMQ consistently improves over naive PMQ baselines for Simple Averaging, Task Arithmetic, and TIES-Merging, and remains competitive for the stronger WUDI merged model.

\subsection{Full 14-Task Results}
\label{app:clip_14task_full}

\tabClipBThirtyTwoFourteenTaskFullPmq

Table~\ref{tab:clip-b32-14task-full-pmq} reports the full 14-task CLIP-ViT-B/32 results.
E-PMQ substantially improves over RTN, GPTQ, and AWQ, especially for weaker upstream mergers such as Task Arithmetic.

\tabClipLFourteenFourteenTaskFullPmq

Table~\ref{tab:clip-l14-14task-full-pmq} reports the full 14-task CLIP-ViT-L/14 results.
The gains remain consistent on the larger CLIP backbone, showing that the improvement is not specific to CLIP-ViT-B/32.

\subsection{Full 20-Task Results}
\label{app:clip_20task_full}

\tabClipBThirtyTwoTwentyTaskFullPmq

Table~\ref{tab:clip-b32-20task-full-pmq} reports the full 20-task CLIP-ViT-B/32 results.
This setting is more challenging for naive PMQ, and E-PMQ provides large improvements by using expert-guided targets during quantization.

\tabClipLFourteenTwentyTaskFullPmq

Table~\ref{tab:clip-l14-20task-full-pmq} reports the full 20-task CLIP-ViT-L/14 results.
E-PMQ again outperforms naive PMQ baselines across the main merging methods, confirming its scalability to larger task suites.

\section{LLM implementation details.}
\label{app:implementation_details_llm}
For the Llama experiments, we construct merged models using Task Arithmetic with
the merge coefficient set to \(0.3\). The source experts correspond to four
capability-oriented tasks: instruction following, coding, mathematics, and
multilingual understanding. We evaluate Llama-3.1-3B and Llama-3.1-8B merged
models.

All quantized LLMs use 4-bit weight-only quantization. We set the group size to
32, the calibration batch size to 4, and the maximum calibration sequence length
to 512. Each method uses 256 calibration samples. The saved quantized weights
use bfloat16 as the storage dtype for non-quantized tensors. We compare E-PMQ
with AWQ, and GPTQ, and also report the full-precision merged model without
quantization. For AWQ, we set the grid-search parameter to 20 in the reported
baseline experiments. E-PMQ uses the same calibration data and follows the same
forward-order layer-wise quantization protocol as the other PMQ methods.

For E-PMQ, the expert model paths are instantiated from the corresponding model
family and task name. We set the global anchor scaling hyperparameter to
\(\alpha=1\) for the LLM experiments. Source experts are used only during
calibration to construct expert-guided targets. After quantization, the deployed
model is a single 4-bit merged model without additional experts or
inference-time modules.

\section{Full Anchor-Strength Ablation}
\label{app:alpha_ablation_full}

\tabAlphaAblationFull

Table~\ref{tab:alpha_ablation_full} reports the full per-task anchor-strength ablation.
Removing the anchor by setting \(\alpha=0\) causes severe collapse, while positive anchor strengths remain stable and consistently outperform GPTQ in average accuracy.
This supports the necessity of merged-weight anchoring in the E-PMQ objective.

\section{Implementation Details}
\label{app:implementation_details}

Unless otherwise stated, all quantized models use weight-only quantization with group size 128.
The calibration batch size is 32 and the evaluation batch size is 128.
The main experiments use 256 calibration samples per task; for a \(K\)-task merged model, this corresponds to \(256K\) total calibration samples.

For E-PMQ, \(X_i^\ell\) is collected from the current quantization trajectory of the partially quantized merged model on the calibration subset associated with expert/task \(i\).
The expert-guided target is constructed as \(W_i^\ell X_i^\ell\).
No expert-specific activation trajectory is used.

We set \(\alpha=0.01\) for Simple Averaging, Task Arithmetic, and TIES-Merging, and \(\alpha=10\) for WUDI-Merging.
For bit-width analysis, we keep the same calibration protocol and vary only the weight bit-width.
For calibration-budget analysis, we vary the number of calibration samples per task while keeping the remaining quantization settings fixed.
Test sets are used only for final evaluation.

% 中：强烈鼓励匿名补充 ZIP 中附可运行代码供审稿；录用后按政策去匿名链接。审稿人须对材料保密。
% EN: Strongly encouraged to attach reviewable anonymized code; de-anonymize after acceptance as required. Reviewers must keep materials confidential.
\newpage
% 中：Checklist 为投稿强制部分，缺则可能 desk reject；提交前删除 checklist.tex 中 BEGIN/END INSTRUCTIONS 英文说明段，保留标题与题目。
% EN: Checklist is mandatory (desk reject if missing). Before submission, delete the instruction block in \texttt{checklist.tex}, keep the heading and items.
% \input{checklist.tex}

\end{document}